%% file: main.tex
\documentclass{artificer}
\usepackage{marvosym}

\input{math_commands.tex}

\usepackage{hyperref}
\usepackage{url}
\usepackage{amsmath}
\usepackage{amssymb}

\usepackage{graphicx}
\usepackage{booktabs}
\usepackage{multirow}
\usepackage{makecell}
\usepackage{subcaption}
\usepackage{wrapfig}
\usepackage{float}

\usepackage{algorithm}
\usepackage{algpseudocode}

\usepackage{xcolor}
\usepackage{microtype}
\newcommand{\algbox}[2][1]{%
  \parbox[t]{\dimexpr\linewidth-(\algorithmicindent)*#1\relax}{\raggedright #2}}

\newcommand{\modelname}{EDEN}

\title{Beyond Policy Rollouts: Embedding-Space Directional Exploration with Anchored Noise for LLM RLVR}

\author{Daijiao Liu}
\author{Huiyi Wang}
\author{Lina Yao}
\author[\textnormal{\Letter}]{Dong Gong}
\affiliation{University of New South Wales (UNSW Sydney)}
\emails{\{daijiao.liu, huiyi.wang, lina.yao, dong.gong\}@unsw.edu.au}
\contribution{\textsuperscript{\Letter}Corresponding author.}

\abstract{
Reinforcement learning with verifiable rewards (RLVR) typically uses small rollout groups per prompt, where repeated sampling from the same policy can yield highly similar trajectories, reducing within-group reward variation and weakening group-relative advantage signals.  
Existing remedies such as higher sampling temperature, entropy bonuses, asymmetric clipping, and larger rollout budgets broaden sampling or modify policy updates but leave rollout conditioning unchanged. In this paper, we propose \textbf{\modelname} (\emph{Embedding-space Directional Exploration via anchored Noise}), which intervenes at the \emph{rollout source} through anchored input perturbations. 
\modelname{} explores beyond the policy's own rollouts. Within the same rollout budget, it replaces a small fraction of standard rollouts in each group with rollouts that the same policy generates from the prompt plus a single appended, perturbed embedding position, rather than from the prompt alone. Gaussian noise drives variation, while a shared anchor provides a common conditioning direction within each prompt group. Selected perturbed rollouts contribute to GRPO updates with downweighted advantages, and the model architecture, discrete decoding, and test-time inference remain unchanged.
Across model scales and mathematical reasoning benchmarks, \modelname{} outperforms GRPO and the exploration controls evaluated in the same pipeline, with the largest gains in 
sampled solution coverage.
Analyses show that more prompt groups provide non-zero group-relative learning signals and that noise and anchoring play complementary roles. Intervening at the rollout source is thus an effective lever for exploration in RLVR.
}

\begin{document}
\maketitle

\section{Introduction}
\label{sec:intro}
Reinforcement learning with verifiable rewards (RLVR) has become the standard recipe for post-training reasoning LLMs, instantiated in algorithms such as PPO~\citep{schulman2017proximal}, RLOO~\citep{ahmadian2024back}, GRPO~\citep{shao2024deepseekmath,guo2025deepseek}, and DAPO~\citep{yu2025dapo}. Most of them share a group-rollout backbone: for each prompt, the policy samples $G$ trajectories whose verifiable rewards drive the update, making rollout quality a direct constraint on learning. Rollout generation is expensive, so typical training uses small groups of 8–16 rollouts per prompt. At this budget, repeated sampling from the same policy can yield highly similar trajectories, and the gradient sees a narrow slice of the solution space~\citep{wachi2026relative}. The key constraint is therefore not rollout count alone, but the amount of useful variation within the rollout pool.

This rollout homogeneity can cause the group-relative advantage signal that GRPO~\citep{shao2024deepseekmath} relies on to collapse. When all rollouts in a group receive the same reward, their advantages vanish, motivating dynamic-sampling filters in recent RLVR systems~\citep{yu2025dapo}. Over training, policy entropy can collapse within a few hundred steps~\citep{cui2025entropymechanismreinforcementlearning}, and pass@$k$ can flatten or even regress below the base model for large $k$~\citep{yue2026does,wen2026reinforcement,yang2025depth,wachi2026relative}. These observations highlight exploration as a shared bottleneck: the rollout pool needs more diversity than the converging policy alone can supply~\citep{zhu2025soft}, making exploration a central research focus in RLVR.

\begin{wrapfigure}{R}{0.54\columnwidth}
    \centering
    \includegraphics[width=\linewidth]{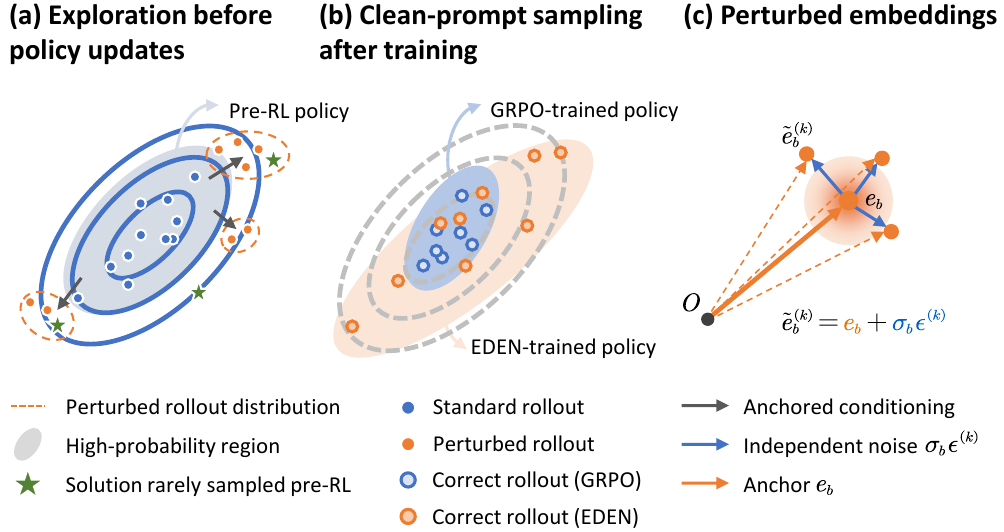}
    \caption{\modelname{} explores beyond the policy's rollouts. (a)~Anchored conditioning yields rollouts from a shifted proposal. (b)~After training, \modelname{} achieves broader coverage of correct trajectories under clean-prompt sampling. (c)~Independent Gaussian noise perturbs a shared anchor in embedding space.
    }
    \label{fig:concept}
\end{wrapfigure}
Existing exploration methods for RLVR address this through different leverage points. Higher sampling temperatures~\citep{dang2026temperature,zhang2025making} broaden token-level sampling by flattening the output distribution, while entropy bonuses, asymmetric clipping (Clip-Higher)~\citep{yu2025dapo}, and entropy-collapse regularizers such as Clip-Cov / KL-Cov~\citep{cui2025entropymechanismreinforcementlearning} aim to prevent premature concentration or relax policy updates. These methods encourage exploration by modifying the training objective or the sampling distribution, while leaving rollout conditioning unchanged. Other methods apply \emph{selection}~\citep{lin2025cppo,zheng2025act,yu2025dapo}, pruning low-information trajectories or skipping groups with uniform rewards. A distinct set of approaches directly alters rollout generation. Search-based methods branch from intermediate states~\citep{xing2026lookahead}, whereas perturbation-based methods modify soft-token representations, policy parameters, and logits, or repurpose the noise inherent in quantised execution~\citep{butt2026soft,bai2026psn,huang2025qerl,zheng2025softgrpo}.
These works suggest that the locus of exploration matters, since changing the conditions under which trajectories are generated can alter the effective trajectory distribution beyond simply drawing more rollouts from the same distribution. 
Can a minimal input-embedding intervention enable useful trajectory-level exploration while retaining standard discrete decoding and leaving test-time inference unchanged?

\begin{figure}[t]
    \centering
    \includegraphics[width=0.95\linewidth]{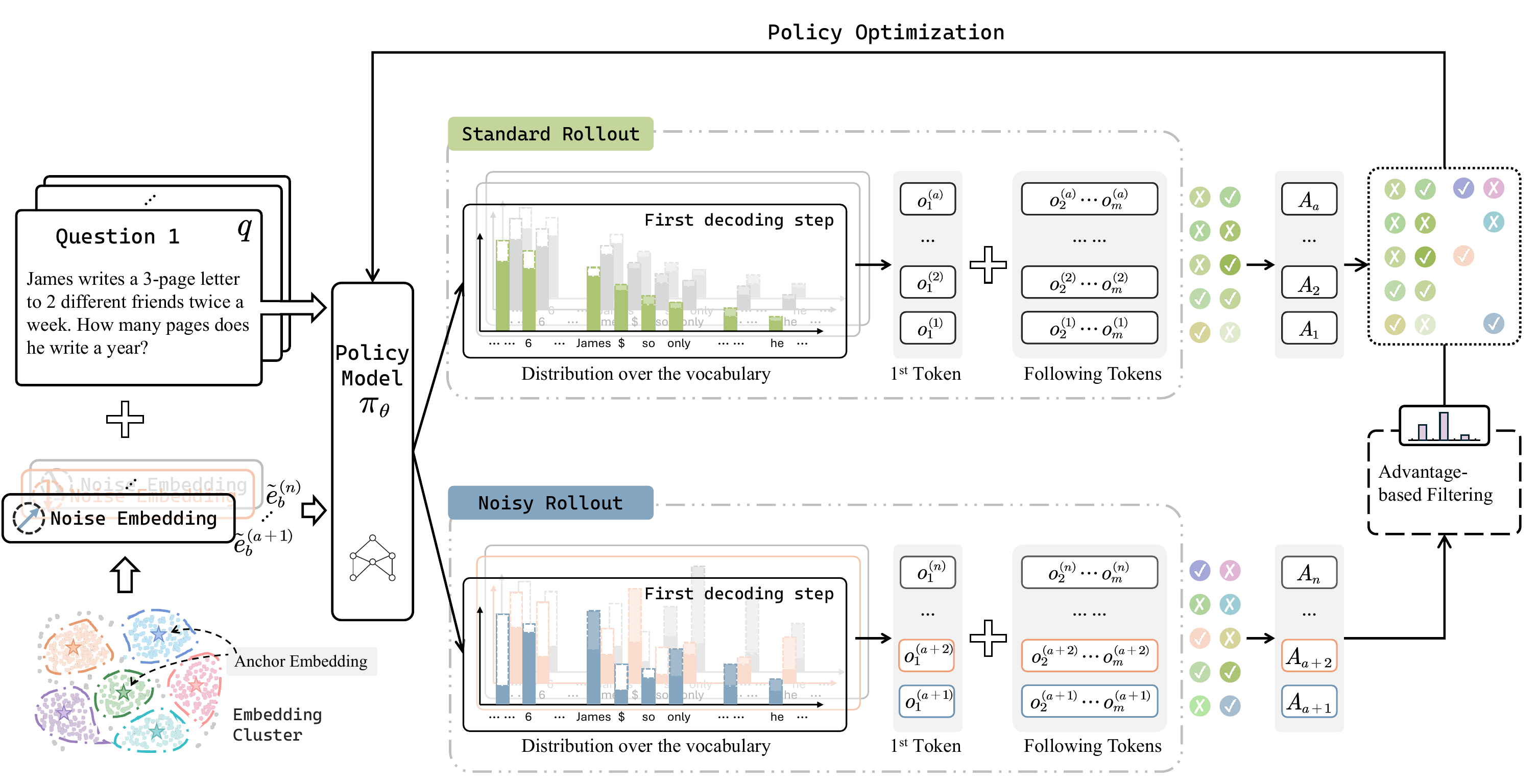}
    \caption{Overview of EDEN. EDEN augments RLVR with a small fraction of noise-perturbed rollouts to broaden trajectory-level exploration without increasing the total rollout budget.}
    \label{fig:model_diagram}
\end{figure}

In this paper, we propose \textbf{\modelname} (\emph{Embedding-space Directional Exploration via anchored Noise}), an RLVR exploration method that intervenes at the rollout source through anchored input perturbations. Without additional rollouts, \modelname{} explores beyond the policy's  own rollouts (Fig.~\ref{fig:concept}) by conditioning the same policy on  perturbed inputs for a subset of each group, helping bring trajectories rarely obtained through standard sampling into training. Applied only during training, this conditioning is introduced at a single embedding position immediately after the prompt and before the first generated token, allowing it to influence the entire autoregressive rollout from the outset without repeatedly intervening during decoding. \modelname{} therefore requires neither policy-parameter perturbations nor specialized decoding, and leaves test-time inference unchanged. The perturbation combines Gaussian noise, which allows rollouts to deviate from the standard generation mode, with an anchor embedding shared by all perturbed rollouts for a prompt. The shared anchor provides directionality by defining a common embedding-space conditioning vector for all perturbed rollouts of a prompt, while independent isotropic Gaussian noise introduces local variation around it. 
To turn this additional exploration into useful training signals, \modelname{} retains perturbed trajectories with large absolute advantages and incorporates them into GRPO updates with downweighted advantages. Fig.~\ref{fig:model_diagram} illustrates the training procedure. EDEN improves over GRPO across three model scales and also outperforms the exploration controls evaluated in the same pipeline. Mechanism analyses further support the effectiveness of input-side exploration. Our main contributions are summarized as follows:
\begin{itemize}
  \item We propose \textbf{\modelname}, a new RLVR exploration method that intervenes at the rollout source through \emph{anchored noise}: a small fraction of rollouts are perturbed via anchored Gaussian noise, and the most informative perturbed rollouts are incorporated into training via an advantage-gated filter and noise-aware advantage reweighting.
  \item \modelname{} realises \emph{directional} exploration through the complementary roles of noise and anchoring, with Gaussian noise driving variation and a chosen anchor embedding supplying a shared conditioning direction within each prompt group. The design integrates with GRPO through a single appended embedding position, retains discrete decoding, and leaves the model architecture and inference unchanged.
  \item Across three model scales and eight mathematical reasoning benchmarks, \modelname{} improves average Pass@1 and Pass@8 over GRPO, with its advantage persisting under Pass@256 evaluation on AIME. Training dynamics show more prompt groups with nonzero training signals, while controlled ablations support the complementary roles of noise and anchoring.
\end{itemize}

\section{Related Work}
\label{sec:related}

\paragraph{RLVR and GRPO foundations.}
RLVR trains reasoning LLMs with verifiable rewards~\citep{lightman2023let,cobbe2021training}, using algorithms such as PPO~\citep{schulman2017proximal}, RLOO~\citep{ahmadian2024back}, and GRPO~\citep{shao2024deepseekmath,guo2025deepseek}. GRPO replaces PPO's learned value function with a group-relative advantage signal computed across rollouts sampled per prompt, and has become the dominant choice for verifiable-reward LLM post-training. DAPO~\citep{yu2025dapo} extends GRPO with asymmetric clipping (Clip-Higher), Dynamic Sampling, token-level loss, and overlong reward shaping, which together stabilise large-scale training. Further variants refine advantage estimation, importance ratios, or training schedules while retaining the group-rollout backbone. Our work builds directly on GRPO and is compatible with these refinements.

\paragraph{Exploration approaches in RLVR.}
Exploration is a central challenge in RLVR, with policy concentration and limited solution coverage motivating a range of approaches~\citep{yue2026does,wen2026reinforcement,yang2025depth,wachi2026relative}. Higher sampling temperatures~\citep{dang2026temperature,zhang2025making}, entropy bonuses, asymmetric clipping~\citep{yu2025dapo}, and Clip-Cov / KL-Cov~\citep{cui2025entropymechanismreinforcementlearning} broaden sampling or counter premature policy concentration. ProRL~\citep{liu2025prorl} extends training with KL resets to sustain exploration. Other methods reshape the optimisation \emph{objective} itself. Pass-at-$k$ Training~\citep{chen2025pass}, FlowRL~\citep{zhu2025flowrl}, and Diversity-Aware Policy Optimization~\citep{yao2025diversity} adjust the reward objective to maintain diversity. These approaches modify sampling or optimization while leaving input conditioning unchanged. \modelname{} instead intervenes at the \emph{rollout source} through anchored input perturbations.

\paragraph{Rollout methods for LLM RLVR.}
Some rollout methods improve efficiency by selecting informative rollouts or prompts, as in Dynamic Sampling~\citep{yu2025dapo}, GRESO~\citep{zheng2025act}, CPPO~\citep{lin2025cppo}, and related approaches~\citep{hu2025vade,qu2026can,tang2026towards,nguyen2026adaptive}, or by reusing historical rollouts to amortise generation cost~\citep{huang2026pros,zhang2025rlep,zhang2025improving,sun2025improving}. Other methods promote exploration by intervening before or during generation. Tree search~\citep{xing2026lookahead} branches from intermediate states to diversify trajectories. Soft Tokens~\citep{butt2026soft} perturbs input embeddings at each CoT step with soft-token decoding, PSN-GRPO~\citep{bai2026psn} perturbs policy weights with truncated importance sampling, QeRL~\citep{huang2025qerl} exploits quantisation-induced noise, SofT-GRPO~\citep{zheng2025softgrpo} applies Gumbel perturbations to logits with soft decoding, and REFT diversifies discrete first-token choices~\citep{kim2026rolloutsbeginlowloadhighleverage}. \modelname{} instead uses a shared anchor to direct continuous perturbations at one appended input embedding for a subset of rollouts.

\section{Method}
\label{sec:methodology}
\suppressfloats[t]

\subsection{Preliminaries}
\label{sec:prelim}
RLVR is widely used for post-training on tasks with automatically checkable solutions such as mathematical reasoning. Concretely, given a prompt $q \sim P(Q)$, the policy $\pi_\theta$ generates a response $o$ by autoregressive sampling and receives a verifiable reward. We build on Group Relative Policy Optimization (GRPO)~\citep{shao2024deepseekmath}, a critic-free policy-gradient algorithm that replaces PPO's learned value function with a group-relative baseline. For each prompt $q$, we sample a group of $G$ rollouts $\{o_i\}_{i=1}^G$ from the old policy $\pi_{\theta_{\mathrm{old}}}$ at sampling temperature $\tau_r$ and optimize
\begin{equation}
\small
\mathcal{J}_{\mathrm{GRPO}}(\theta)
=
\mathbb{E}_{q,\,\{o_i\}}
\!\left[
\frac{1}{G}\sum_{i=1}^{G}\frac{1}{|o_i|}\sum_{t=1}^{|o_i|}
\Big(
\ell^{\mathrm{clip}}_{i,t}(\theta)
-
\beta \,\hat{\mathbb{D}}_{\mathrm{KL}}^{(i,t)}[\pi_\theta \,\|\, \pi_{\mathrm{ref}}]
\Big)
\right],
\label{eq:grpo}
\end{equation}
where $\ell^{\mathrm{clip}}_{i,t}(\theta)
=
\min\!\bigl(
r_{i,t}A_{i,t},\;
\operatorname{clip}(r_{i,t},1-\epsilon,1+\epsilon)A_{i,t}
\bigr)$ and $
r_{i,t}
=
\frac{\pi_\theta(o_{i,t}\mid q,o_{i,<t})}
{\pi_{\theta_{\mathrm{old}}}(o_{i,t}\mid q,o_{i,<t})}$,
$\epsilon$ controls the clipping range. The Kullback--Leibler regularization term against a frozen reference policy $\pi_{\mathrm{ref}}$ is estimated by the non-negative unbiased form
\begin{equation}
\small
\hat{\mathbb{D}}_{\mathrm{KL}}^{(i,t)}[\pi_\theta \,\|\, \pi_{\mathrm{ref}}]
=
\frac{\pi_{\mathrm{ref}}(o_{i,t}\mid q,o_{i,<t})}
{\pi_\theta(o_{i,t}\mid q,o_{i,<t})} - \log \frac{\pi_{\mathrm{ref}}(o_{i,t}\mid q,o_{i,<t})}
{\pi_\theta(o_{i,t}\mid q,o_{i,<t})} - 1.
\label{eq:kl}
\end{equation}
The token-level advantage $A_{i,t} = (r_i - \mu_q)/\sigma_q$ standardizes the per-rollout rewards within the group, where $\mu_q$ and $\sigma_q$ denote the mean and standard deviation of $\{r_j\}_{j=1}^G$. The same scalar advantage is shared by all tokens of $o_i$ when computing the token-level loss.

\subsection{Embedding-space Directional Exploration via anchored Noise}
\label{sec:noise_oriented_optimization}
EDEN reshapes the rollout distribution through controlled embedding-space perturbations. It has three parts. \emph{Perturb} samples a fraction of each group from the same policy under perturbed input conditioning. \emph{Regulate} controls exploration through a shared vocabulary-token anchor, a norm-relative noise scale, a screened anchor bank, and a temperature schedule. \emph{Absorb} incorporates informative perturbed trajectories into updates of the unperturbed policy through advantage-based selection, advantage scaling, and evaluation under the clean prompt.

\subsubsection{Noise-Perturbed Rollout Generation}
\label{sec:noise_perturbed_rollout_generation}
We generate noise-perturbed rollouts by applying scaled noise to an anchor embedding drawn from a pre-constructed bank. Let $E(q) = [\mathbf{e}_1, \mathbf{e}_2, \dots, \mathbf{e}_L]$ denote the prompt embedding sequence. For each prompt, we append one embedding position to the prompt sequence and set it to a perturbed embedding $\tilde{\mathbf{e}}_b$, yielding
$\tilde{E}(q) = \bigl[\mathbf{e}_1, \mathbf{e}_2, \ldots, \mathbf{e}_L, \tilde{\mathbf{e}}_b\bigr]$. Rollouts are then generated as $o \sim \pi_\theta(\cdot \mid \tilde{E}(q))$. This single-position intervention can influence the entire subsequent rollout through autoregressive conditioning, without further perturbations during decoding.

\paragraph{Noise bank construction.}
To give each prompt group's perturbed rollouts a shared conditioning centre while allowing exploration across different embedding-space directions, EDEN draws anchors from a fixed noise bank $\mathcal{B}$ of $K$ vocabulary-token embeddings. The bank is constructed once before RL training and remains fixed throughout training. Anchors are drawn from the model's own vocabulary and screened for generation viability and complementary directions, aiming to induce non-trivial variation without substantially disrupting generation. Appendix~\ref{appedix:noise_bank_construction} details the construction and screening procedure using 200 training-split prompts. A single fixed anchor already improves over GRPO, with the bank providing further gains mainly in greedy Pass@1 (Tables~\ref{tab:random_noise_vs_bank} and~\ref{tab:bank_vs_single}). For each prompt in each training batch, one anchor is sampled uniformly from $\mathcal{B}$ and shared across its $K_n$ perturbed rollouts. Independent anchor sampling across prompts and batches allows exploration around different conditioning centres rather than a single fixed anchor.

\paragraph{Embedding-Space Perturbation.}
Given the shared anchor embedding $\mathbf{e}_b \in \mathbb{R}^d$, we independently sample Gaussian noise $\boldsymbol{\epsilon} \sim \mathcal{N}(\mathbf{0}, \mathbf{I})$ for each perturbed rollout and construct $\tilde{\mathbf{e}}_b = \mathbf{e}_b + \rho \frac{\|\mathbf{e}_b\|_2}{\sqrt{d}} \boldsymbol{\epsilon}$, where $\rho$ controls the relative perturbation scale. This yields per-dimension standard deviation $\sigma_b = \rho \|\mathbf{e}_b\|_2 / \sqrt{d}$ and root-mean-square perturbation magnitude $\sqrt{\mathbb{E}[\|\tilde{\mathbf{e}}_b - \mathbf{e}_b\|_2^2]} = \rho \|\mathbf{e}_b\|_2$. Thus, $\rho$ scales perturbations relative to the anchor norm, stabilizing magnitudes across anchors and model sizes. The Gaussian component is isotropic and has no preferred direction. In high dimensions, the scaled noise is nearly orthogonal to $\mathbf{e}_b$, with norm close to $\rho\|\mathbf{e}_b\|_2$, so the perturbed embedding forms an angle of approximately $\arctan\rho$ with the anchor. \emph{Directional} therefore refers to two properties of the anchor. Within a prompt group, all $K_n$ perturbed rollouts share one anchor and hence one displacement of the conditioning, with local variation from Gaussian noise (Table~\ref{tab:random_noise_vs_bank}). Distinct anchor directions in the bank allow groups to explore different displacements (Table~\ref{tab:bank_vs_single}).

\paragraph{Exploration Control.}
We control exploration strength using a temperature schedule $\tau_n(s)$ that increases from $\tau_0$ to $\tau^*$ over training,
$\tau_n(s) = \tau_0 + (\tau^* - \tau_0)\frac{1}{2}\bigl(1 - \cos(\pi \cdot \mathrm{clip}(s/T_w, 0, 1))\bigr)$, where $s$ is the training step and $T_w$ is the warmup duration.
This schedule limits early randomness for stability while enabling stronger exploration as the policy becomes more reliable.

\subsubsection{Noise-Aware Trajectory Selection and Advantage Estimation}
\label{sec:Noise_aware_trajectory_selection}
For each prompt $q_i$, we form a mixed group of $K_r$ standard and $K_n$ noise-perturbed rollouts, compute their rewards, and normalize advantages over $\mathcal{S}_{r,i} \cup \widetilde{\mathcal{S}}_{n,i}$ as in GRPO. Noisy rollouts are then selected once per prompt batch $\mathcal{P}$, rather than per prompt. From the batch-wide noisy rollout pool $\widetilde{\mathcal{S}}_{n}^{\mathcal{P}} = \bigcup_{q_i \in \mathcal{P}} \widetilde{\mathcal{S}}_{n,i}$, EDEN retains
$\mathcal{S}_{n}^{\mathcal{P}} = \{s \in \widetilde{\mathcal{S}}_{n}^{\mathcal{P}} \mid |A_s| \text{ is among the top fraction } p \text{ of } \widetilde{\mathcal{S}}_{n}^{\mathcal{P}}\}$,
where $p$ is the noisy-rollout keep ratio. Group-wise normalization makes $|A_s|$ comparable across prompts, while batch-level selection allows each prompt to retain a variable number of noisy rollouts rather than a fixed quota. This criterion prioritizes perturbed trajectories that are most informative relative to their prompt groups. Under binary rewards, a non-degenerate group with $G$ rollouts and $m$ correct responses assigns advantages $\sqrt{(G-m)/m}$ to correct rollouts and $-\sqrt{m/(G-m)}$ to incorrect ones. The minority outcome thus has the largest magnitude, so batch-level $|A|$ selection favors rare within-group deviations that change the group-relative signal rather than reinforce the dominant outcome. Fully degenerate all-correct or all-incorrect groups are handled by the same numerical stabilizer used in GRPO normalization.
\begin{wrapfigure}{r}{0.55\textwidth}
\vspace{-\intextsep}
\begin{minipage}{0.55\textwidth}
\begin{algorithm}[H]
\caption{EDEN Training}
\label{alg:noir}
\small
\begin{algorithmic}[1]
\Require policy $\pi_\theta$, noise bank $\mathcal{B}$, temperatures $\tau_r$, $\tau_n(s)$,
rollout counts $K_r, K_n$, keep ratio $p$, advantage scale $\gamma$

\For{$s = 1, \dots, S$}
    \State Sample a prompt batch $\mathcal{P} = \{x_i\}_{i=1}^{B}$
    \For{each prompt $x_i \in \mathcal{P}$}
        \State Sample standard rollouts
        $\mathcal{S}_{r,i} \sim \pi_\theta(\cdot \mid x_i; \tau_r)$
        \State Sample anchor $\mathbf{e}_b \sim \mathrm{Uniform}(\mathcal{B})$
        \State \algbox[2]{Sample noisy rollouts
        $\widetilde{\mathcal{S}}_{n,i}
        = \{o^{(k)} \sim
        \pi_\theta(\cdot \mid x_i, \tilde{\mathbf{e}}_b^{(k)}; \tau_n(s))
        \}_{k=1}^{K_n}$}
        \State \algbox[2]{Compute rewards and group-relative advantages on
        $\mathcal{S}_{r,i} \cup \widetilde{\mathcal{S}}_{n,i}$}
    \EndFor
    \State \algbox[1]{$\mathcal{S}_{n}^{\mathcal{P}}
    \gets \mathrm{Top}_{p}\!\left(
    \bigcup_i \widetilde{\mathcal{S}}_{n,i};\, |A| \right)$}
    \State \algbox[1]{$\mathcal{D} \gets
    \{(o,A(o)) : o \in \bigcup_i \mathcal{S}_{r,i}\}
    \cup \{(o,\gamma A(o)) : o \in \mathcal{S}_{n}^{\mathcal{P}}\}$}
    \State Update $\theta$ on $\mathcal{D}$ via Eq.~\ref{eq:noir_loss}
\EndFor
\end{algorithmic}
\end{algorithm}
\end{minipage}
\vspace{-1cm}
\end{wrapfigure}

To control the influence of perturbed trajectories, we scale their advantages as $\tilde{A}_i = \gamma A_i$, where $\gamma \in (0,1]$, while retaining $\tilde{A}_i = A_i$ for standard rollouts. Since perturbed rollouts are optimized under clean prompt conditioning, this scaling reduces their gradient contribution to account for the difference between generation and update conditions.

\subsubsection{Noise-Aware Policy Update}
\label{sec:noise_aware_policy_update}
EDEN updates the policy using standard and selected noise-perturbed rollouts. During the update, the appended perturbation position is removed, and token log-probabilities for all trajectories are evaluated under the clean prompt embedding $E(q)$. Thus, selected noisy trajectories are generated from $\pi_\theta(\cdot \mid \tilde{E}(q))$ but optimized under $\pi_\theta(\cdot \mid E(q))$. EDEN does not introduce an importance-sampling correction for this generation-update mismatch. Instead, advantage filtering determines which perturbed trajectories enter the update, and advantage scaling by $\gamma$ reduces their gradient contribution. We use the following GRPO-style objective:
\begin{equation}
\begin{aligned}
\mathcal{J}_{\mathrm{EDEN}}(\theta)
&=
\mathbb{E}_{q}
\!\left[
\frac{1}{|\mathcal{S}(q)|}\sum_{i \in \mathcal{S}(q)}
\frac{1}{|o_i|}\sum_{t=1}^{|o_i|}
\Big(
\min\!\bigl(
r_{i,t}\tilde{A}_i,\;
\operatorname{clip}\!\left(r_{i,t},1-\epsilon_{\mathrm{low}},1+\epsilon_{\mathrm{high}}\right)\tilde{A}_i
\bigr)
\Big)
\right] \\
&\quad
- \beta \,\mathbb{D}_{\mathrm{KL}}(\pi_\theta \,\|\, \pi_{\mathrm{ref}})
+ \alpha \,\mathbb{H}(\pi_\theta),
\end{aligned}
\label{eq:noir_loss}
\end{equation}
where $\mathcal{S}(q)$ denotes the set of rollouts used for training for prompt $q$, and $\tilde{A}_i$ is the scaled advantage. The detailed training procedure is summarized in Algorithm~\ref{alg:noir}.

\section{Experiments}
\label{sec:experiments}

\textbf{Models, Datasets, and Baselines.}
We evaluate on three Qwen2.5 models at different scales: Qwen2.5-0.5B~\citep{qwen2025qwen25technicalreport}, Qwen2.5-Math-1.5B~\citep{yang2024qwen2}, and Qwen2.5-Math-7B~\citep{yang2024qwen2}. For Qwen2.5-0.5B, we train on GSM8K~\citep{cobbe2021training} and evaluate on GSM8K and MATH~\citep{hendrycks2021measuring}. For the Qwen2.5-Math models, we train on MATH and DAPO-Math~\citep{yu2025dapo}, and evaluate on GSM8K, MATH, AMC, AIME 2024, AIME 2025, OlympiadBench~\citep{he2024olympiadbench}, Gaokao~\citep{zhang2023evaluating}, and MinervaMath~\citep{lewkowycz2022solving}. Beyond GRPO, we compare with the exploration controls in the same pipeline, none of which changes the rollout input conditioning: a higher sampling temperature, a larger entropy bonus, asymmetric clipping, and larger rollout groups.

\textbf{Training Details.} 
All experiments are conducted in the verl framework with vLLM for rollout generation. We use a binary reward, assigning $1$ to correct responses and $0$ otherwise. 
Each prompt produces $16$ rollouts by default. For EDEN, these include $12$ standard and $4$ noise-perturbed rollouts. The noise bank size is 8. $\rho$ is set to $0.12$ for Qwen2.5-0.5B and $0.18$ for the other models. We train with a batch size of $128$ and a learning rate of $1\times 10^{-6}$. We set $\beta=0$ and $\alpha=0.001$. All methods use this standard setting, and Table~\ref{tab:seed_robustness_05b} reports seed robustness on Qwen2.5-0.5B. We train Qwen2.5-0.5B for $6$ epochs and Qwen2.5-Math-1.5B and Qwen2.5-Math-7B for $10$ epochs. For Pass@8 and Pass@256 evaluations, we set the sampling temperature to $0.6$ and top-$p$ to $0.95$. All experiments are run on $8\times$H100 GPUs. Additional implementation details are provided in Appendix~\ref{sec:appendix_setup}.

\begin{table}[t]
\centering
\caption{
Performance on mathematical reasoning benchmarks with Qwen2.5-Math-1.5B. We report Pass@1 and Pass@8. \textbf{Bold} indicates the best result per column.
}
\label{tab:qwen1.5b_results}
\small
\setlength{\tabcolsep}{4pt}
\renewcommand{\arraystretch}{1.1}
\begin{tabular}{l l cccccc c}
\toprule
Metric & Method & MATH & GSM8K & AMC & Olympiad & Gaokao & Minerva & Avg. \\
\midrule
\multirow{4}{*}{Pass@1} 
 & GRPO          & 76.60 & \textbf{86.35} & 53.01 & 40.50 & 64.94 & \textbf{31.99} & 58.90 \\
 & + Higher Temp. & 77.00 & 85.29 & 56.63 & 39.02 & 65.71 & 30.15 & 58.97 \\
 & + Clip Higher & \textbf{77.40} & 84.69 & 56.63 & 41.25 & 64.16 & 31.62 & 59.29 \\
 & EDEN    & \textbf{77.40} & 86.28 & \textbf{63.86} & \textbf{41.84} & \textbf{67.01} & 31.62 & \textbf{61.34} \\
\midrule
\multirow{4}{*}{Pass@8} 
 & GRPO          & 86.60 & 94.92 & 69.88 & 55.34 & 77.14 & 41.54 & 70.90 \\
 & + Higher Temp. & 88.20 & 91.89 & 71.08 & 53.26 & 77.40 & 42.28 & 70.69 \\
 & + Clip Higher & 87.40 & 95.07 & 74.70 & 54.90 & 76.88 & 42.28 & 71.87 \\
 & EDEN    & \textbf{89.60} & \textbf{95.60} & \textbf{79.52} & \textbf{55.93} & \textbf{78.96} & \textbf{44.12} & \textbf{73.96} \\
\bottomrule
\end{tabular}
\end{table}

\begin{table}[t]
\centering
\caption{
Performance on mathematical reasoning benchmarks with Qwen2.5-Math-7B. We report Pass@1 and Pass@8. \textbf{Bold} indicates the best result per column.
}
\label{tab:qwen7b_results}
\small
\setlength{\tabcolsep}{3pt}
\renewcommand{\arraystretch}{1.1}
\begin{tabular}{llccccccc}
\toprule
Metric & Method & MATH & GSM8K & AMC & Olympiad & Gaokao & Minerva & Avg. \\
\midrule
\multirow{2}{*}{Pass@1} 
 & GRPO        & \textbf{82.00} & 88.93 & \textbf{71.08} & 44.81 & 66.75 & 35.66 & 64.87 \\
 & EDEN  & 81.80 & \textbf{90.30} & 68.67 & \textbf{47.48} & \textbf{68.57} & \textbf{36.03} & \textbf{65.48} \\
\midrule
\multirow{2}{*}{Pass@8} 
 & GRPO        & 88.80 & 96.21 & \textbf{83.13} & 58.31 & 77.14 & 42.65 & 74.37 \\
 & EDEN  & \textbf{93.00} & \textbf{97.04} & \textbf{83.13} & \textbf{64.39} & \textbf{80.52} & \textbf{50.00} & \textbf{78.01} \\
\bottomrule
\end{tabular}
\end{table}

\subsection{Main Results}
\label{sec:exp-main}
EDEN outperforms GRPO on average at both 1.5B and 7B scales across English and Chinese mathematics benchmarks from grade-school to olympiad level. On Qwen2.5-Math-1.5B, we also compare with same-pipeline exploration controls, including \textbf{Higher Temperature} ($T=1.2$) for increased sampling randomness and \textbf{Clip Higher}~\citep{yu2025dapo} for asymmetric PPO clipping. The entropy bonus and larger rollout groups are examined in Section~\ref{sec:analysis}. EDEN achieves the best average Pass@1 and Pass@8 in Table~\ref{tab:qwen1.5b_results}, exceeding GRPO by $2.44$ and $3.06$ points, respectively. Neither control alone yields comparable gains, highlighting the benefit of EDEN's exploration strategy. The improvement persists at 7B (Table~\ref{tab:qwen7b_results}). Across both scales, the larger gains on Pass@8 than Pass@1 are consistent with improved coverage of valid solutions under a fixed sampling budget.

\subsection{Analysis}
\label{sec:analysis}
\paragraph{Exploration on Hard Problems}
\label{sec:exp-aime-pass256}
To assess solution coverage at a larger sampling budget, we evaluate Pass@256 on the challenging AIME'24 and AIME'25 benchmarks. EDEN improves average Pass@256 over GRPO by 6.67 points across model scales, including a 13.33 point gain on AIME'25 on Qwen2.5-Math-7B (Table~\ref{tab:aime_results}). These results suggest that EDEN's solution-coverage gains persist even at a large evaluation budget. Fig.~\ref{fig:aime_grop_vs_noir} shows that after training EDEN produces more correct AIME rollouts, with more of them from problems solved by at least four distinct approaches and fewer from problems solved by a single approach. The case studies in Appendix~\ref{appendix:case_studies} provide qualitative examples of how this diversity appears in complete reasoning traces.

\begin{table*}
\centering

\begin{minipage}[t]{0.53\textwidth}
\centering
\caption{Pass@256 on AIME benchmarks. EDEN's advantage persists at high sampling budget. }
\label{tab:aime_results}
\small
\setlength{\tabcolsep}{4pt}

\resizebox{\linewidth}{!}{%
\begin{tabular}{llccc}
\toprule
\textbf{Model} & \textbf{Method}
& \makecell{\textbf{AIME2024}\\[-1pt]\footnotesize Pass@256}
& \makecell{\textbf{AIME2025}\\[-1pt]\footnotesize Pass@256}
& \textbf{Avg.} \\
\midrule
\multirow{2}{*}{Qwen2.5-Math-1.5B}
& GRPO & 53.33 & 50.00 & 51.67 \\
& EDEN & \textbf{60.00} & \textbf{56.67} & \textbf{58.33} \\
\midrule
\multirow{2}{*}{Qwen2.5-Math-7B}
& GRPO & \textbf{73.33} & 53.33 & 63.33 \\
& EDEN & \textbf{73.33} & \textbf{66.67} & \textbf{70.00} \\
\bottomrule
\end{tabular}
}
\end{minipage}
\hfill
\begin{minipage}[t]{0.4\textwidth}
\centering
\vspace{0pt}
\includegraphics[width=\linewidth]{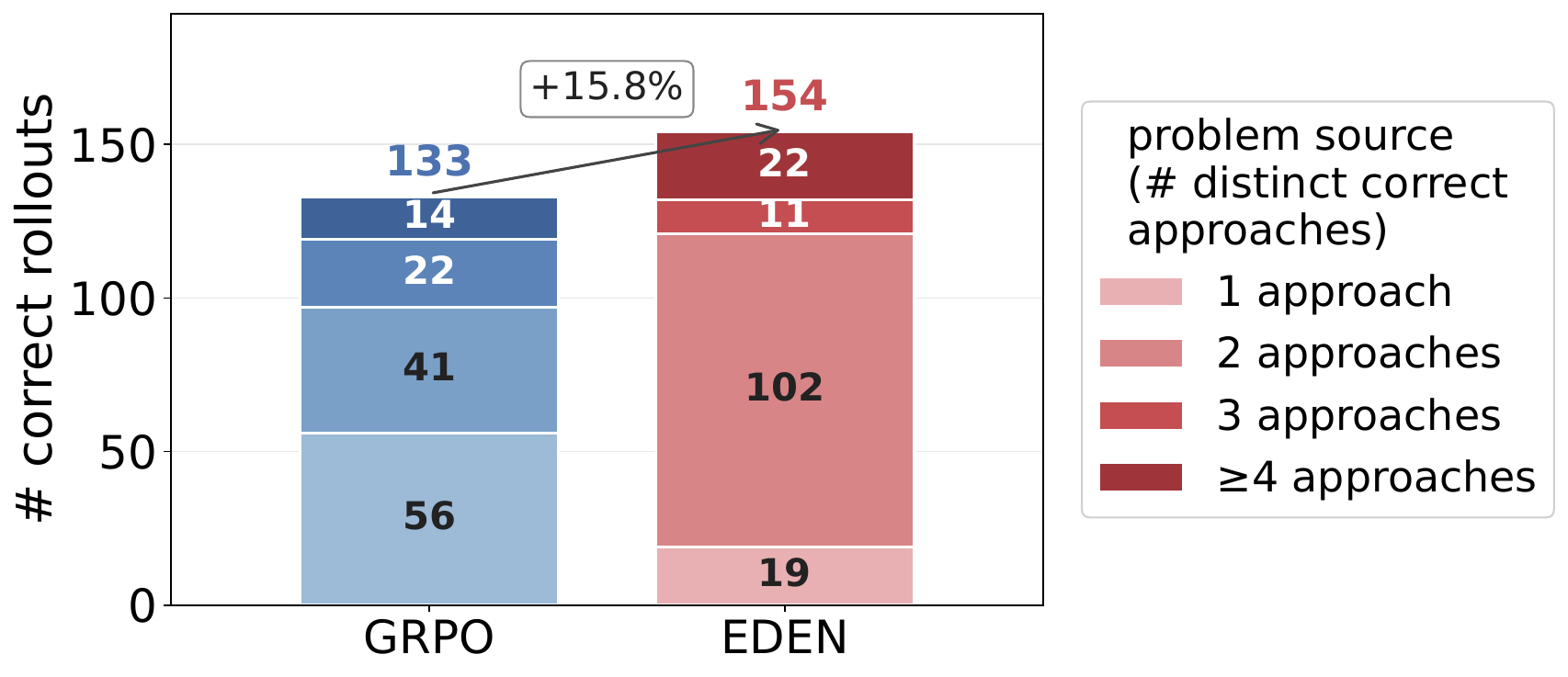}

\captionsetup{
    type=figure,
    font=footnotesize,
    justification=raggedright,
    singlelinecheck=false
}
\vspace{-0.1cm}
\captionof{figure}{Correct AIME 2024/2025 rollouts (16 per problem, 60 problems), grouped by per-problem reasoning-pattern counts.}
\label{fig:aime_grop_vs_noir}
\end{minipage}
\end{table*}

\begin{figure}[t]
    \centering
    \begin{minipage}[t]{0.32\linewidth}
        \centering
        \includegraphics[width=\linewidth]{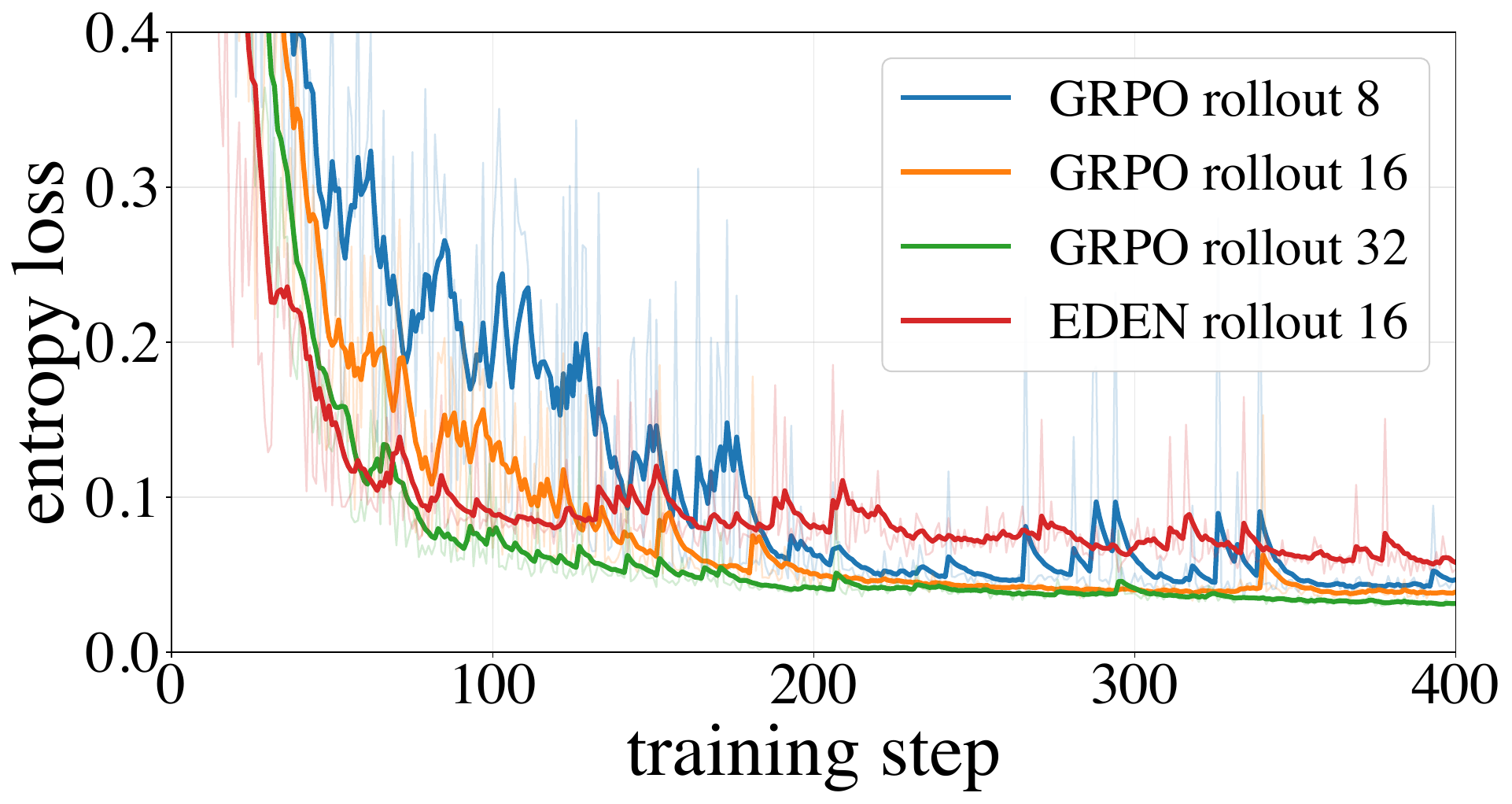}
        \captionof{figure}{
            Training entropy on Qwen2.5-0.5B.
        }
        \label{fig:entropy_loss_0_5b}
    \end{minipage}
    \hfill
    \begin{minipage}[t]{0.63\linewidth}
        \centering
        \begin{minipage}[t]{0.48\linewidth}
            \centering
            \includegraphics[width=\linewidth]{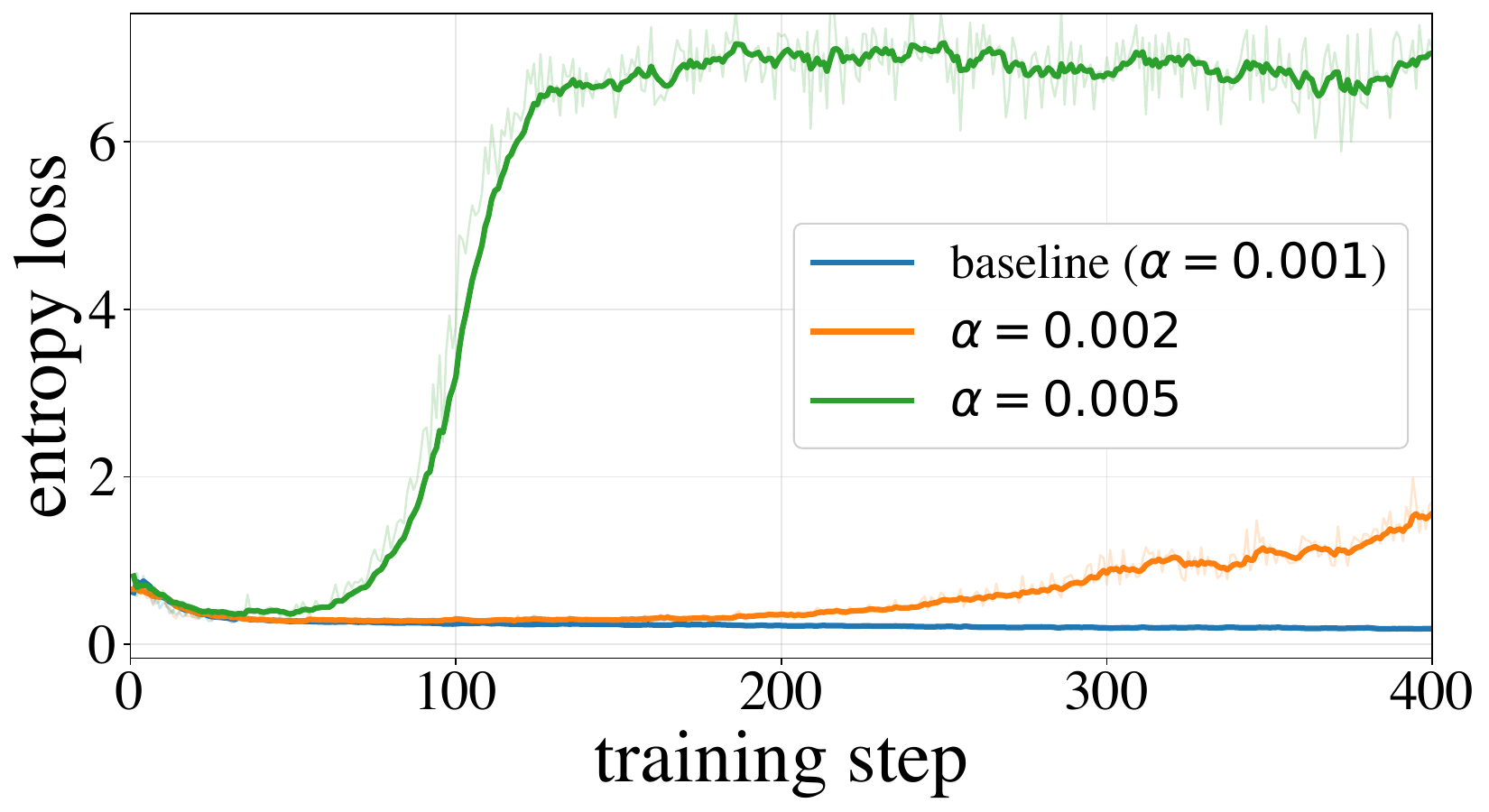}
        \end{minipage}
        \hfill
        \begin{minipage}[t]{0.48\linewidth}
            \centering
            \includegraphics[width=\linewidth]{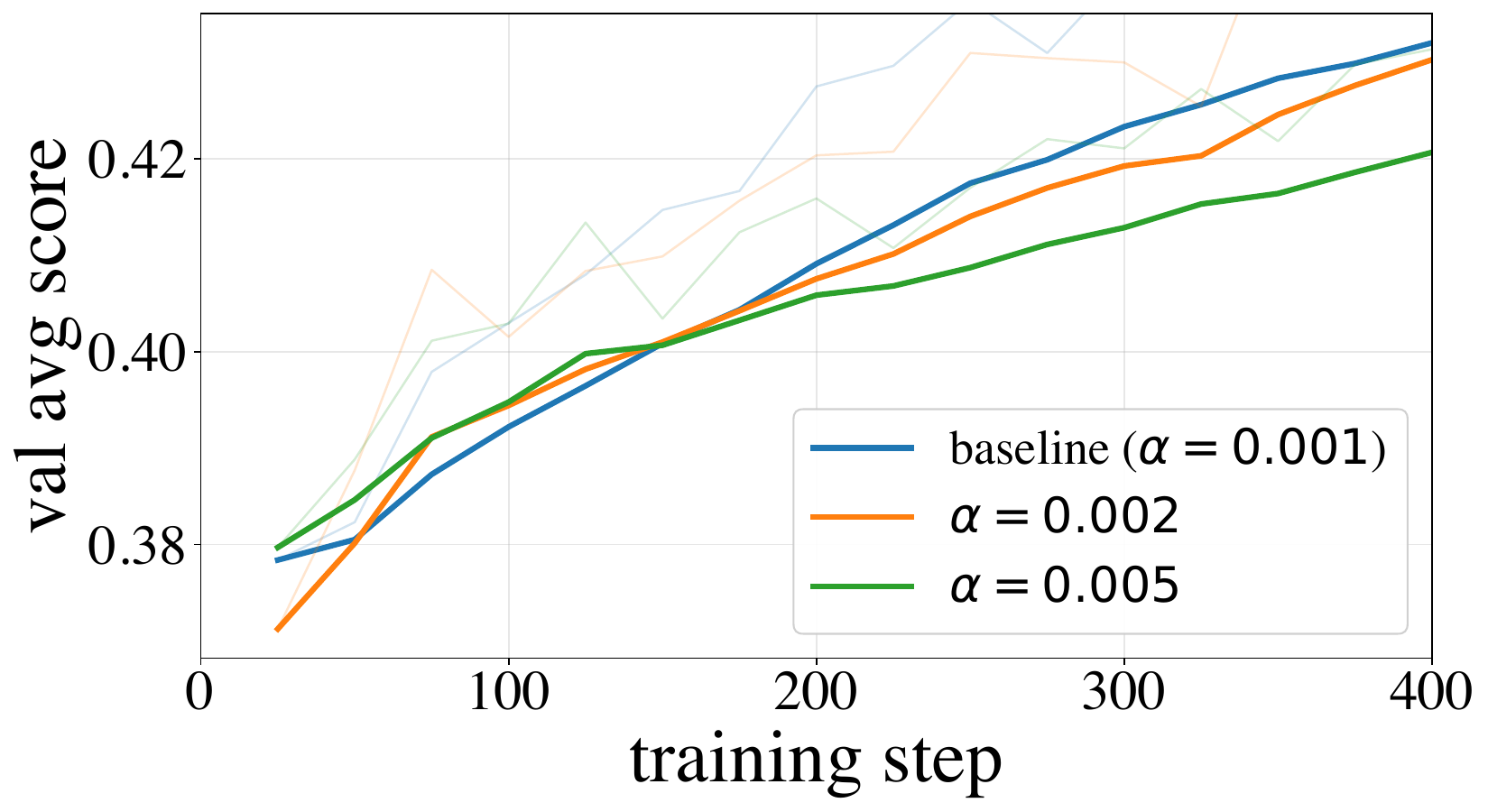}
        \end{minipage}

        \captionof{figure}{
            Effect of the entropy bonus coefficient.
            Left: Policy entropy. Right: Validation Avg@4.
        }
        \label{fig:entropy_bonus_ablation}
    \end{minipage}
\end{figure}

\paragraph{Breaking the Rollout Budget Ceiling}
A simple alternative to EDEN's structured perturbation is to increase GRPO's rollout budget, potentially providing more gradient signal and broader sampling. We compare GRPO with 8, 16, and 32 rollouts per prompt against EDEN with 16. Increasing GRPO's budget from 8 to 32 improves Pass@1 by only 0.5 points and slightly reduces Pass@8, whereas EDEN substantially outperforms GRPO at both matched and larger GRPO budgets (Table~\ref{tab:rollout_pass}). Fig.~\ref{fig:entropy_loss_0_5b} shows that all GRPO configurations reach similarly low entropy by step 200, suggesting that sustaining exploration requires more than drawing additional samples from the same distribution. EDEN maintains higher entropy than every GRPO configuration after warm-up. Together, these results suggest that increasing GRPO's rollout count alone does not reproduce EDEN's sustained exploration or performance gains.

\paragraph{Role of Anchored Noise.}
Table~\ref{tab:random_noise_vs_bank} examines how noise, anchor choice, and anchor sharing affect EDEN. Removing noise reduces performance to the level of GRPO, showing that an unperturbed anchor alone is insufficient. Norm-matched random noise without an anchor also fails to reproduce EDEN's gains. A fixed random anchor recovers part of the improvement but remains below EDEN, showing that consistent anchoring can be useful even without bank diversity. Selecting a new random anchor for every rollout performs worse, which is consistent with increased within-group dispersion when rollouts from the same prompt use different conditioning centers. EDEN instead shares one curated-bank anchor within each prompt group and samples anchors across groups, combining a common within-group conditioning centre with variation across groups.

\begin{table}[t]
\centering
\footnotesize
\setlength{\tabcolsep}{6pt}
\caption{Qwen2.5-0.5B performance under varying perturbations and anchor construction, averaged over MATH500 and GSM8K (16 rollouts/prompt). $^{\dagger}$ Three anchor draws.}
\label{tab:random_noise_vs_bank}
\begin{tabular}{llcc}
\toprule
\textbf{Method} & \textbf{Construction} & \textbf{Pass@1} & \textbf{Pass@8} \\
\midrule
GRPO & --- & 45.67 & 66.36 \\
\midrule
\modelname{} & curated bank ($K{=}8$), per prompt & \textbf{47.95} & \textbf{69.01} \\
\quad -- w/o noise & curated bank ($\rho=0$), no perturbation & 45.42 & 66.51\\
\quad -- w/o anchor & norm-matched random noise & 43.39 & 65.45 \\
\quad -- random anchor & random vocab anchor, per rollout & 43.86 & 65.81 \\
\quad -- random fixed anchor$^{\dagger}$ & one random vocab anchor, fixed & 46.37 & 67.78 \\
\bottomrule
\end{tabular}
\end{table}

\paragraph{Comparison with Exploration Controls}
We examine whether EDEN's gains can be explained by conventional exploration controls applied during generation or policy updates. Increasing the entropy bonus on Qwen2.5-Math-1.5B raises policy entropy throughout training without improving validation accuracy, which declines at larger bonuses (Fig.~\ref{fig:entropy_bonus_ablation}). Higher sampling temperatures yield inconsistent gains, and extreme temperatures destabilize training under the default settings (Appendix~\ref{appendix:ablation_studeis}). Clip Higher~\citep{yu2025dapo} provides limited and inconsistent gains over GRPO, whereas EDEN achieves stronger performance with the same asymmetric clipping, supporting the benefit of its additional rollout-side intervention. These results suggest that simply broadening the policy distribution does not reproduce EDEN's gains, supporting its use of structured perturbations.

\paragraph{Why EDEN Works: Expanding the Training Signal}
\label{sec:noir_work_reason}
EDEN affects both trajectory diversity and group-relative training signals. Even before RL training, correct anchor-perturbed trajectories spread more widely than standard rollouts in aligned 4-gram space (Fig.~\ref{fig:aligned_correct_scatter_by_method}), with an RMS radius about 1.6 times as large and more samples outside the standard-rollout hull. Anchored noise thus changes the surface form of correct trajectories before any policy update. Visualization details are provided in Appendix~\ref{appendix:trajectory_diversity}.
Fig.~\ref{fig:non_diverse_ratio} shows fewer zero-advantage groups under EDEN from about step 750 onward, reaching $0.53$ versus $0.58$ for GRPO at step 2000. Unlike post-hoc filtering of a fixed rollout distribution, EDEN changes the distribution itself, allowing otherwise uniform-reward groups to yield mixed outcomes and provide non-zero group-relative training signals. 

\subsection{Ablation Studies}
\label{sec:ablation}
\paragraph{Effect of Advantage Filtering.}
We study the role of advantage-based filtering in EDEN by comparing against a variant that removes this mechanism and uses all noisy rollouts during training. As shown in Fig.~\ref{fig:adv_filter_combined}, removing the filter leads to unstable updates, with frequent spikes in gradient norms due to high-variance noisy rollouts. In contrast, by selectively retaining trajectories with large absolute advantages, EDEN yields smoother gradient dynamics and consistently better performance across the evaluated benchmarks.

\noindent
\begin{minipage}[htp]{0.56\linewidth}
    \vspace{0pt}
    \centering
    \scriptsize
    \setlength{\tabcolsep}{2.5pt}
    \captionof{table}{Performance on Qwen2.5-0.5B under varying rollout budgets.}
    \begin{tabular}{llcccccc}
    \toprule
    \multirow{2}{*}{Method} & 
    & \multicolumn{3}{c}{Pass@1} 
    & \multicolumn{3}{c}{Pass@8} \\
    \cmidrule(lr){3-5} \cmidrule(lr){6-8}
    & 
    & MATH500 & GSM8K & Avg.
    & MATH500 & GSM8K & Avg. \\
    \midrule
    \multirow{3}{*}{GRPO} 
    & 8  & 31.20 & 59.36 & 45.28 & 59.20 & 76.50 & 67.85 \\
    & 16 & 31.60 & 59.74 & 45.67 & 58.20 & 74.53 & 66.36 \\
    & 32 & 30.00 & \textbf{61.56} & 45.78 & 57.60 & 74.83 & 66.21 \\
    \midrule
    EDEN 
    & 16 
    & \textbf{34.80} & 61.11 & \textbf{47.95}
    & \textbf{59.40} & \textbf{78.62} & \textbf{69.01} \\
    \bottomrule
    \end{tabular}
    \label{tab:rollout_pass}
\end{minipage}
\hspace{0.5cm}
\begin{minipage}[htp]{0.35\linewidth}
    \centering
    \includegraphics[width=\linewidth]{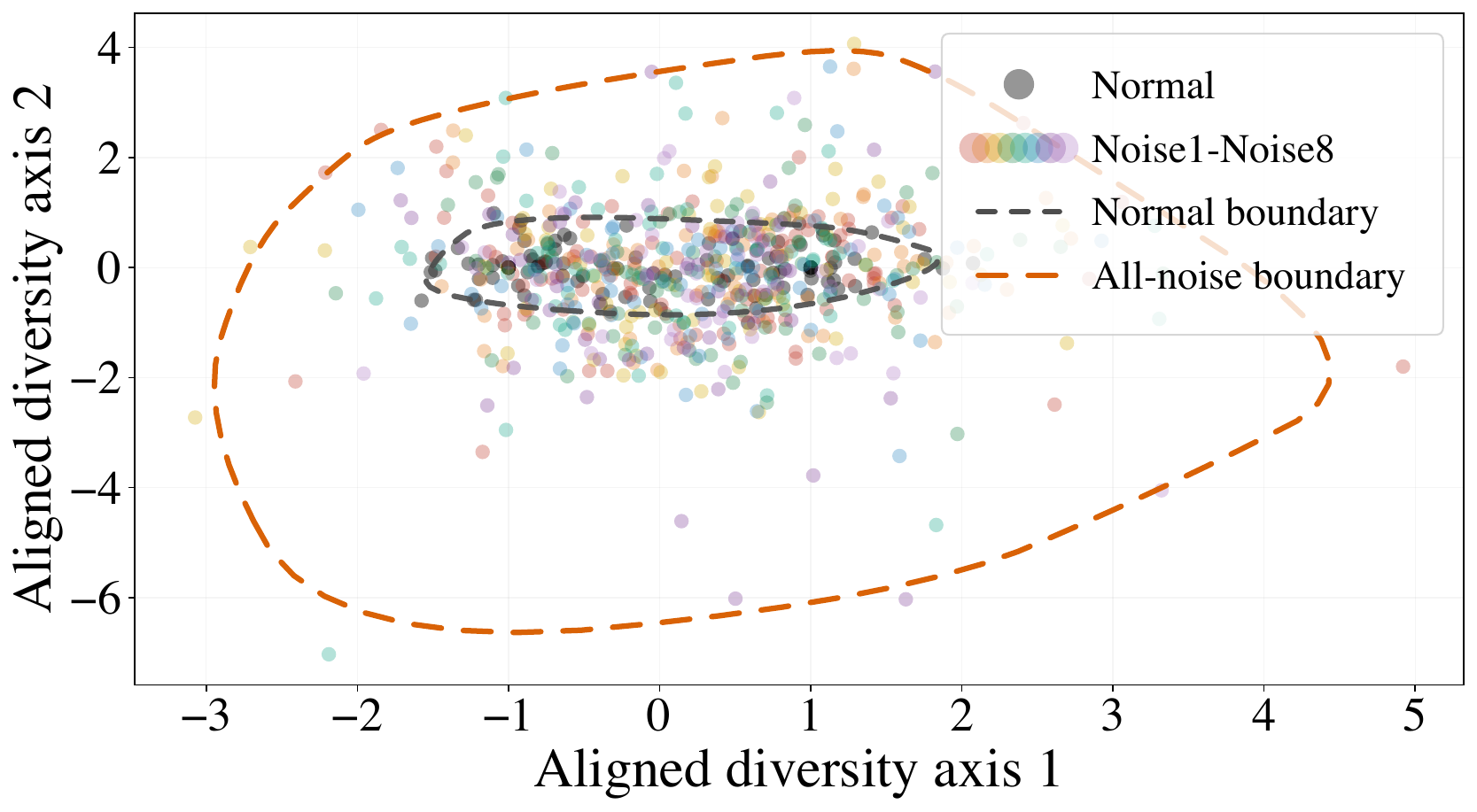}
    \captionof{figure}{Diversity of correct trajectories in aligned 4-gram space.}
    \label{fig:aligned_correct_scatter_by_method}
\end{minipage}

\begin{figure}[t]
    \centering
    \begin{subfigure}[t]{0.32\linewidth}
        \centering
        \includegraphics[width=\linewidth]{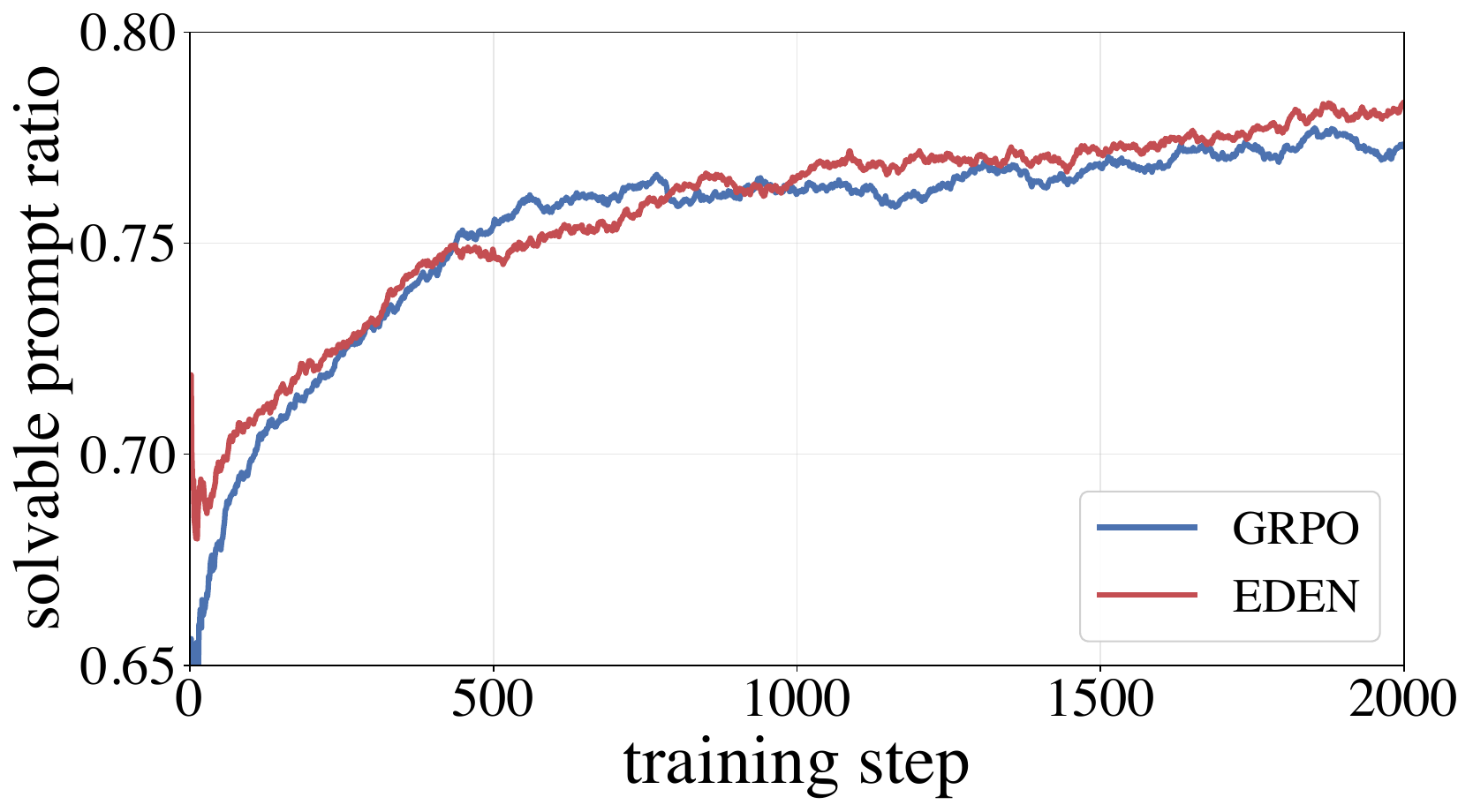}
        \caption{Fraction of prompts with at least one correct rollout.}
        \label{fig:solvable_ratio_2000}
    \end{subfigure}
    \hfill
    \begin{subfigure}[t]{0.32\linewidth}
        \centering
        \includegraphics[width=\linewidth]{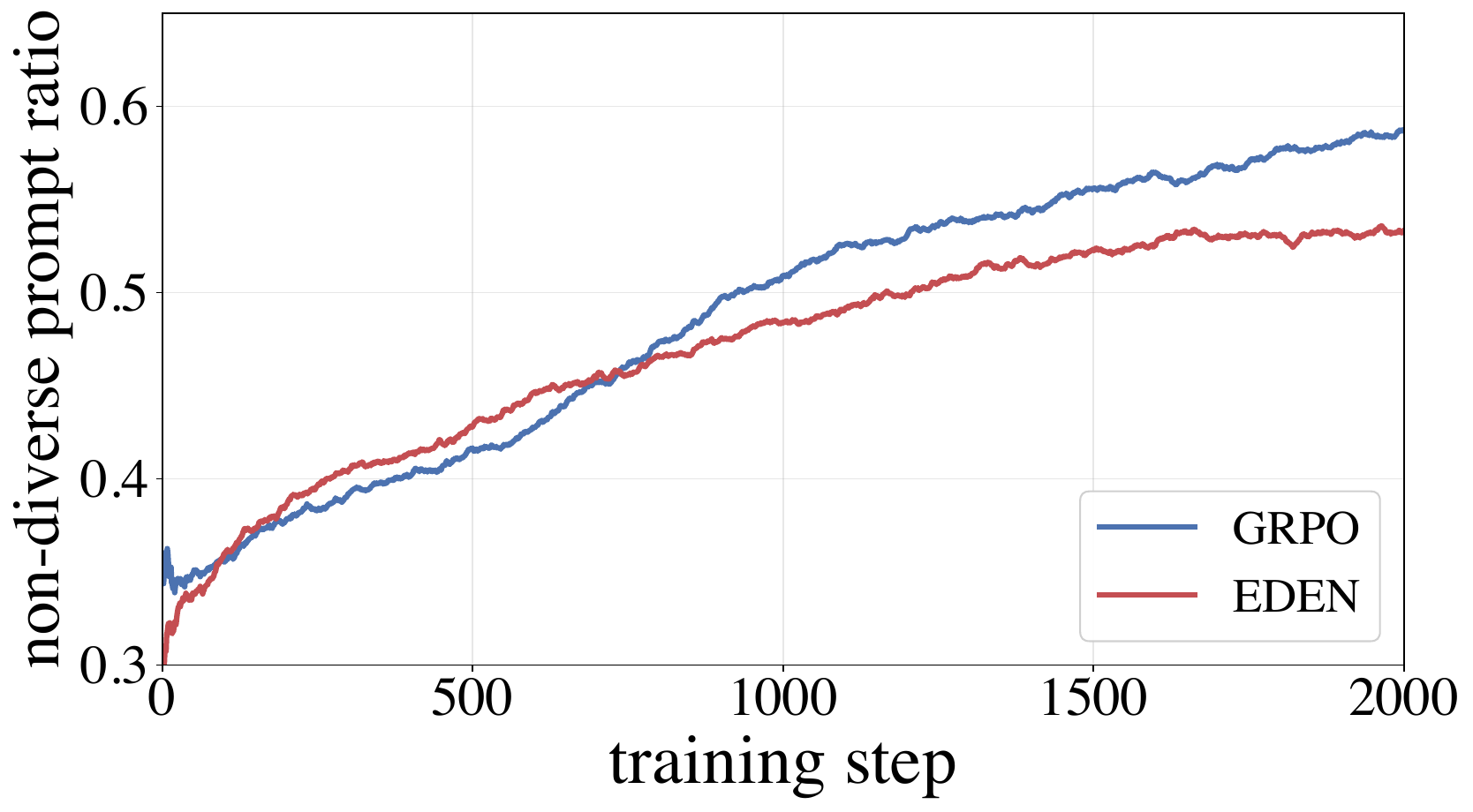}
        \caption{Fraction of prompts with zero advantages (all-correct/all-incorrect).}
        \label{fig:non_diverse_ratio}
    \end{subfigure}
    \hfill
    \begin{subfigure}[t]{0.32\linewidth}
        \centering
        \includegraphics[width=\linewidth]{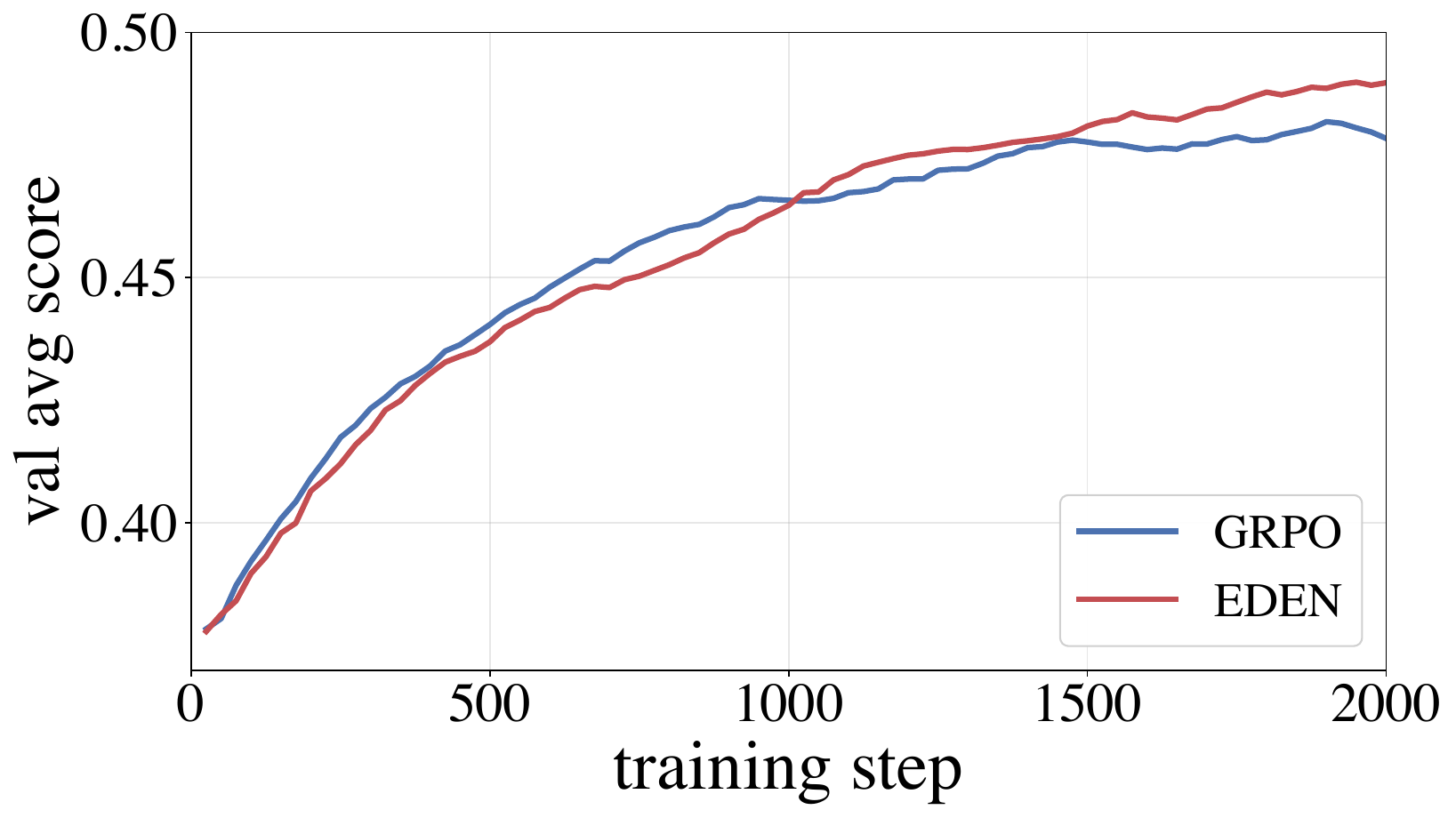}
        \caption{Validation Avg@4 averaged across all eight benchmarks.}
        \label{fig:acc_grpo_noir}
    \end{subfigure}
    \caption{Training dynamics of EDEN vs.\ GRPO on Qwen2.5-Math-1.5B. }
    \label{fig:why_noir_work}
\end{figure}

\begin{figure}[htp]
\centering

\begin{minipage}{0.54\linewidth}
    \centering
    \includegraphics[width=0.75\linewidth]{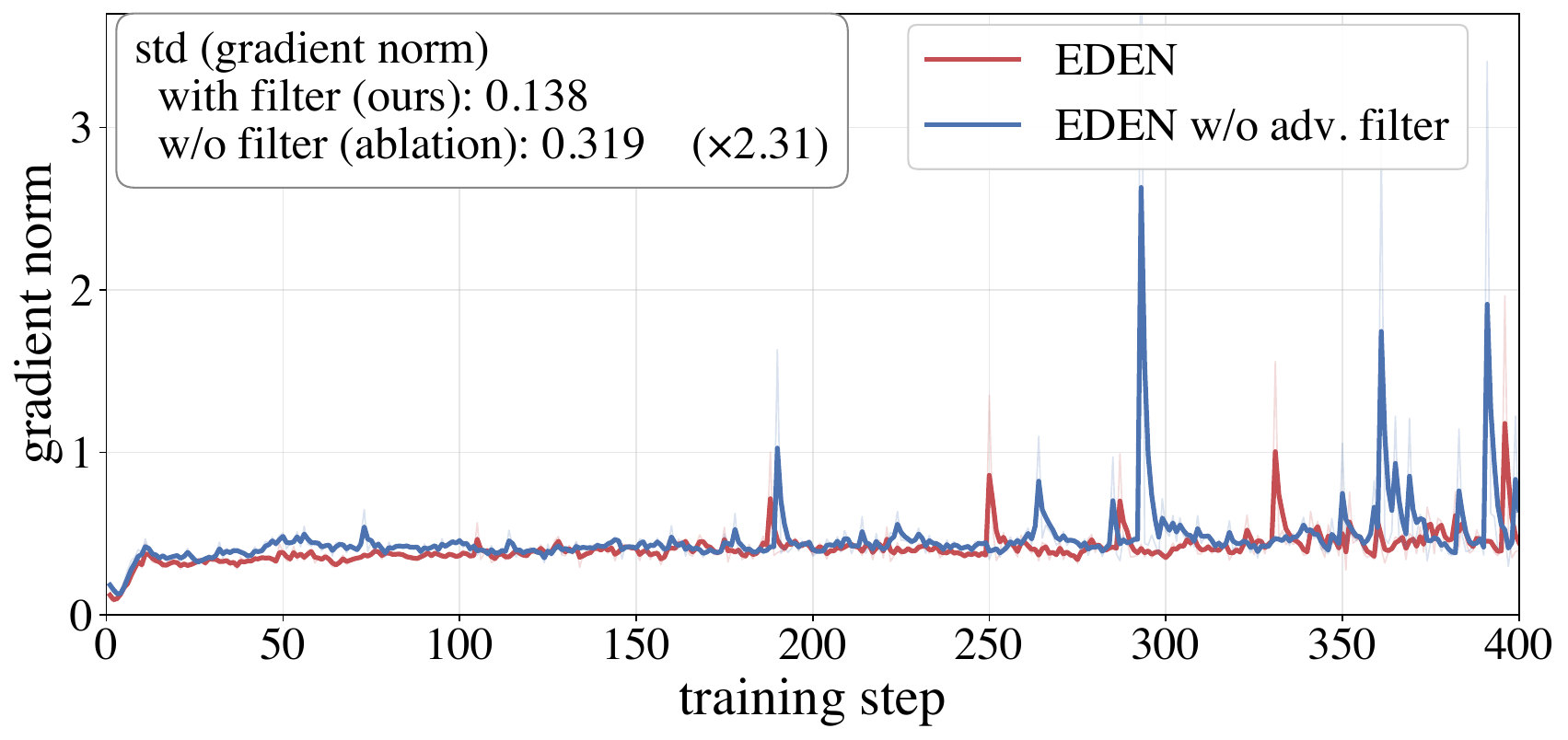}
\end{minipage}
\hfill
\begin{minipage}{0.4\linewidth}
    \centering
    \footnotesize
    \setlength{\tabcolsep}{3pt}
    \begin{tabular}{@{}lccc@{}}
    \toprule
    Method & MATH & GSM8K & Avg. \\
    \midrule
    EDEN & \textbf{34.10} & \textbf{59.95} & \textbf{47.03} \\
    - adv. filter & 31.85 & 59.38 & 45.62 \\
    \bottomrule
    \end{tabular}
\end{minipage}
\caption{
Effect of advantage filtering. Left: training gradient norm. Right: Avg@4 performance.
}
\label{fig:adv_filter_combined}
\end{figure}

\begin{minipage}{0.31\linewidth}
    \centering
    \footnotesize
    \setlength{\tabcolsep}{3pt}
    \captionof{table}{Effect of varying $K$. Results averaged over all benchmarks.}
    \begin{tabular}{lccc}
        \toprule
        \textbf{Method} & $K$ & \textbf{Pass@1} & \textbf{Pass@8} \\
        \midrule
        GRPO  & -- & 48.34 & 59.43 \\
        \midrule
        EDEN & 1 & 48.45 & 61.96 \\
        EDEN & 4 & 48.58 & 62.02 \\
        EDEN & 8 & \textbf{49.33} & \textbf{62.13} \\
        \bottomrule
    \end{tabular}
    \label{tab:bank_vs_single}
\end{minipage}
\hfill
\begin{minipage}{0.32\linewidth}
    \centering
    \includegraphics[width=\linewidth]{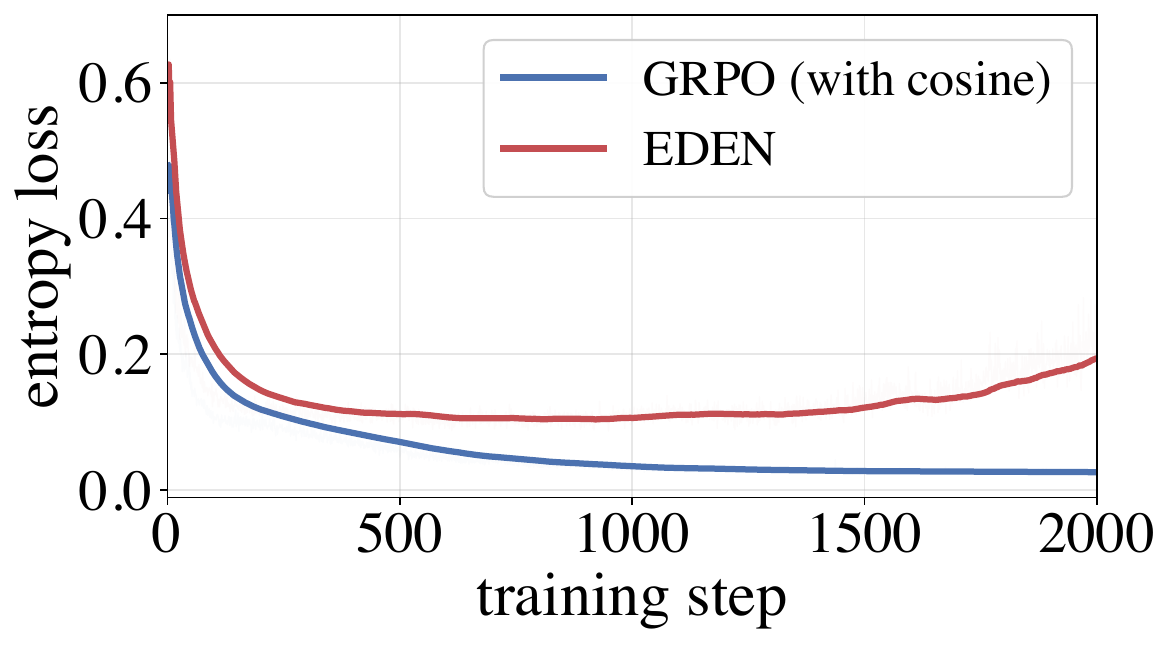}
    \captionof{figure}{
    Qwen2.5-Math-1.5B training entropy with temperature warmup on 4 rollouts.
    }
    \label{fig:entropy_loss_1_5b}
\end{minipage}
\hfill
\begin{minipage}{0.32\linewidth}
    \centering
    \includegraphics[width=\linewidth]{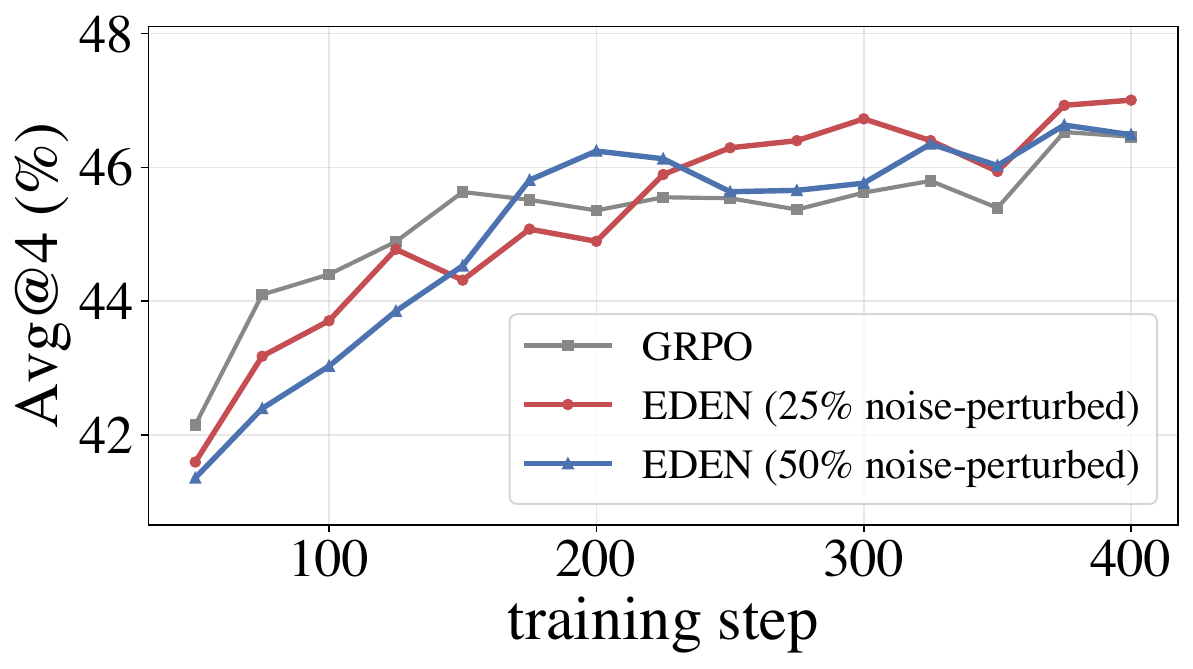}
    \captionof{figure}{
    Effect of noisy rollout fraction on Qwen2.5-0.5B (Avg@4).
    }
    \label{fig:0_5b_16rollouts_comparison}
\end{minipage}

\paragraph{Anchor Bank Size and Noisy Rollout Fraction.}
We compare $K \in \{1,4,8\}$ on Qwen2.5-Math-1.5B in Table~\ref{tab:bank_vs_single}. A single anchor from the constructed bank recovers nearly all of the Pass@8 gain over GRPO, showing that anchored perturbation remains effective in the minimal setting. Larger banks provide modest additional gains, especially on Pass@1, motivating our choice of $K=8$. EDEN thus benefits from complementary anchor centers without requiring a large bank. We also vary the noisy rollout fraction on Qwen2.5-0.5B, keeping 16 rollouts per prompt and comparing 0\% (GRPO), 25\% (12 standard + 4 noisy), and 50\% (8 standard + 8 noisy). The 25\% setting performs best, while 50\% matches GRPO (Fig.~\ref{fig:0_5b_16rollouts_comparison}). A modest noisy fraction thus suffices, while larger fractions leave fewer standard rollouts to provide reliable training signals.

\paragraph{Cosine schedule for exploration control strength.}
EDEN warms up the sampling temperature of noisy rollouts with a cosine schedule, as using the target temperature from the start leads to high entropy and near-zero rewards.  With the same temperature warmup but no embedding perturbations, the baseline's entropy quickly collapses, whereas EDEN maintains higher entropy (Fig.~\ref{fig:entropy_loss_1_5b}). This suggests that the sustained exploration is not explained by the schedule alone. Warmup lengths of 200 and 400 steps yield similar accuracy, indicating limited sensitivity to duration (Appendix~\ref{appendix:ablation_studeis}).

\section{Conclusion}
RLVR faces limited exploration under small rollout budgets, with concentrated trajectories and weak learning signals. We propose \modelname{}, which explores beyond the policy's own rollouts through directional input-embedding perturbations at the rollout source. It generates a fraction of each group under perturbed conditioning and absorbs informative trajectories into the unperturbed policy through selective incorporation and advantage reweighting. This enables structured exploration without changing the model architecture, rollout budget, or test-time inference path. Across model scales and benchmarks, \modelname{} outperforms GRPO and the exploration controls in the same pipeline. These results suggest gains from input-side perturbations beyond output-side exploration controls alone.

\paragraph{Limitations and Future Work.}
Several directions may further extend EDEN's capabilities. First, the current noise bank is constructed from a set of anchor embeddings. More adaptive or learnable constructions, such as dynamic selection or joint optimization with the policy, may enable more targeted exploration. Second, EDEN applies Gaussian perturbations centered at anchor embeddings. More structured schemes, such as anisotropic noise or task-dependent directions, may further improve coverage of the trajectory space.

\bibliography{references}
\bibliographystyle{plainnat}

\clearpage
\appendix

\section{Detailed Experimental Setup}
\label{sec:appendix_setup}
\paragraph{Datasets.}
We evaluate our method across a diverse set of mathematical reasoning datasets, covering a wide range of difficulty levels.

For smaller models (Qwen2.5-0.5B), we train on the GSM8K training set and evaluate on both GSM8K and MATH. GSM8K focuses on grade-school arithmetic reasoning, while MATH contains more challenging competition-style problems.

For math-specialized models (Qwen2.5-Math-1.5B and Qwen2.5-Math-7B), we train on a combination of the MATH training set and DAPO-Math, which together provide diverse and high-difficulty mathematical reasoning data. Evaluation is conducted on a broad suite of benchmarks, including GSM8K and MATH (standard reasoning tasks), AMC, AIME 2024, and AIME 2025 (competition-level problems), as well as OlympiadBench, Gaokao, and MinervaMath, which further extend coverage to olympiad-style and advanced mathematical reasoning tasks.

\paragraph{Models.}
We conduct experiments on three Qwen2.5 models spanning different scales: Qwen2.5-0.5B\footnote{\url{https://huggingface.co/Qwen/Qwen2.5-0.5B}}, Qwen2.5-Math-1.5B\footnote{\url{https://huggingface.co/Qwen/Qwen2.5-Math-1.5B}}, and Qwen2.5-Math-7B\footnote{\url{https://huggingface.co/Qwen/Qwen2.5-Math-7B}}. 

The Qwen2.5-0.5B model serves as a general-purpose base model, while Qwen2.5-Math-1.5B and Qwen2.5-Math-7B are specialized for mathematical reasoning through domain-specific pretraining. This setup allows us to evaluate the effectiveness of our method across both general and math-specialized models, as well as across different model scales.

\paragraph{Training Details.}
All experiments are implemented using the \texttt{verl} framework with vLLM for rollout generation. We use a binary outcome reward that assigns $1$ to correct responses and $0$ otherwise. Across methods, the global batch size is $128$ and the learning rate is $1 \times 10^{-6}$. GRPO uses symmetric PPO clipping with $\epsilon=0.2$. EDEN and the Clip Higher baseline use asymmetric clipping with $\epsilon_{\mathrm{low}}=0.2$ and $\epsilon_{\mathrm{high}}=0.28$. We set the KL coefficient to $0$ and the entropy coefficient to $0.001$. The maximum prompt and response lengths are $1536$ and $2560$ tokens, respectively. Qwen2.5-0.5B is trained for approximately $400$ steps, corresponding to $6$ epochs, while the math-specialized models are trained for approximately $2000$ steps, corresponding to $10$ epochs. All experiments are run on $8\times$H100 GPUs.

GRPO and EDEN use the same generation budget of $16$ rollouts per prompt. GRPO generates all $16$ rollouts from the unmodified input, whereas EDEN generates $12$ standard rollouts and $4$ noise-perturbed rollouts. For each prompt, EDEN samples one anchor from a bank of $8$ candidates and shares it across the $4$ perturbed rollouts, while applying an independent Gaussian perturbation to the shared anchor for each rollout. We set the relative perturbation scale $\rho$ to $0.12$ for Qwen2.5-0.5B and $0.18$ for the math-specialized models.

For GRPO and EDEN, standard rollouts use a fixed sampling temperature of $\tau_r=1.0$. The temperature for noise-perturbed rollouts follows a cosine schedule from $0.2$ to $1.0$ over the first $250$ training steps for Qwen2.5-0.5B, $200$ steps for Qwen2.5-Math-1.5B, and $400$ steps for Qwen2.5-Math-7B, and remains at $1.0$ thereafter. Rewards and group-relative advantages are computed jointly over all $16$ rollouts for each prompt. All standard rollouts are retained for policy updates. We rank the perturbed rollouts globally across each prompt batch by absolute advantage and retain the top $p=0.5$ fraction, so the number retained for an individual prompt can range from $0$ to $K_n$. The advantages of the retained perturbed rollouts are scaled by $\gamma=0.5$. 

\paragraph{Anchor Assignment.}
Sharing one anchor across the perturbed rollouts of a prompt keeps the conditioning direction fixed within the group, while independent Gaussian perturbations introduce local variation around that direction. The anchor is resampled for each prompt, exposing different rollout groups to different directions from the fixed bank.

\section{Full Training Procedure of EDEN}
\label{appendix:algorithm_noir}

We provide the full training procedure of EDEN in Algorithm~\ref{alg:noir_full_alg}. 
Compared to the simplified version presented in the main text, this algorithm explicitly details the construction of mixed rollout batches, including both standard and noise-perturbed rollouts, as well as the subsequent advantage computation and policy updates. In particular, for each prompt, EDEN first samples a set of standard rollouts, followed by a set of noise-anchored rollouts constructed via embedding-space perturbations. These rollouts are combined and filtered based on advantage scores, after which the policy is updated using the selected samples. 

This formulation highlights the integration of structured noise injection with standard policy optimization, illustrating how EDEN balances exploration and stability during training.

\begin{algorithm}[H]
\caption{Full Training Procedure of EDEN}
\label{alg:noir_full_alg}
\begin{algorithmic}[1]
\Require policy $\pi_{\theta}$, noise bank $\mathcal{B}$, rollout temperature $\tau_r$, noise-perturbed temperature schedule $\tau_n(s)$, rollout counts $K_r, K_n$, keep ratio $p$, advantage scale $\gamma$
\For{$s = 1$ to $S$}
\State Sample a prompt batch $\mathcal{P} = \{x_i\}_{i=1}^{B}$
\For{each prompt $x_i \in \mathcal{P}$}
\State Sample standard rollouts $\mathcal{S}_{r,i} = \{o^{(k)}\}_{k=1}^{K_r} \sim \pi_\theta(\cdot \mid x_i; \tau_r)$
\State Sample anchor embedding $\mathbf{e}_b \sim \mathrm{Uniform}(\mathcal{B})$
\For{$k = 1$ to $K_n$}
\State Construct perturbed embedding $\tilde{\mathbf{e}}_b^{(k)}$ from $\mathbf{e}_b$
\State Sample noise-perturbed rollout $o^{(k)} \sim \pi_\theta(\cdot \mid x_i, \tilde{\mathbf{e}}_b^{(k)}; \tau_n(s))$
\EndFor
\State $\widetilde{\mathcal{S}}_{n,i} \gets \{o^{(k)}\}_{k=1}^{K_n}$
\State Compute rewards and group-relative advantages on $\mathcal{S}_{r,i} \cup \widetilde{\mathcal{S}}_{n,i}$
\EndFor
\State $\mathcal{S}_{n}^{\mathcal{P}} \gets \mathrm{Top}_{p}\bigl(\bigcup_i \widetilde{\mathcal{S}}_{n,i};\, |A|\bigr)$
\State $\mathcal{D} \gets \{(o, A(o)) : o \in \bigcup_i \mathcal{S}_{r,i}\} \cup \{(o, \gamma A(o)) : o \in \mathcal{S}_{n}^{\mathcal{P}}\}$
\State Update $\theta$ on $\mathcal{D}$ via Eq.~\ref{eq:noir_loss}
\EndFor
\end{algorithmic}
\end{algorithm}

\section{Additional Details on Noise Bank Construction}
\label{appedix:noise_bank_construction}
The noise bank $\mathcal{B}$ is constructed separately for each model once before RL training and remains fixed throughout policy optimization. Candidate anchors are drawn from the model's own vocabulary and organized as anchor-relative directions in the embedding space. Given a reference token $a$ and another token $i \neq a$, we define
\begin{equation}
\mathbf{u}_i = \frac{\Delta_i}{\|\Delta_i\|_2},
\qquad
\Delta_i = \mathbf{e}_i - \mathbf{e}_a.
\end{equation}
Normalizing the embedding displacement removes the influence of scale and retains the direction from the reference token to each candidate anchor.

We cluster the resulting directions and retain 32 representative candidate anchors for offline screening. For Qwen2.5-0.5B, the screening set contains 200 prompts from the GSM8K training set. For Qwen2.5-Math-1.5B and Qwen2.5-Math-7B, it contains 200 prompts from the DAPO-Math training data. For each candidate and screening prompt, we generate eight perturbed rollouts using the same perturbation construction as in training.

The final bank is selected so that its perturbed rollouts complement the standard rollout distribution. We use the lexicographic objective
\begin{equation}
\mathcal{B}^\star = \arg\max_{\mathcal{B}} \bigl( \mathrm{Cov}(\mathcal{B}),\, \mathbf{1}[\mathrm{LenRatio}(\mathcal{B}) \geq \eta_{\mathrm{len}}],\, \mathrm{DisAns}(\mathcal{B}),\, \mathrm{DisReason}(\mathcal{B}),\, \mathrm{TotCorrect}(\mathcal{B}) \bigr),
\label{eq:bank-objective}
\end{equation}
Here, $\mathrm{Cov}(\mathcal{B})$ is the fraction of prompts with at least one correct perturbed rollout, and $\mathrm{LenRatio}(\mathcal{B})$ enforces response-length feasibility. $\mathrm{DisAns}(\mathcal{B})$ is the mean number of distinct extracted final-answer values per prompt. $\mathrm{DisReason}(\mathcal{B})$ is one minus the mean pairwise Jaccard similarity between token-level 6-gram sets of the normalized reasoning traces, computed within the eight rollouts for each prompt and averaged over the screening prompts; normalization lowercases the traces, removes final-answer patterns, and collapses whitespace. $\mathrm{TotCorrect}(\mathcal{B})$ is the total number of correct rollouts. We prioritize coverage and length feasibility, and use rollout diversity and correctness to rank the remaining candidates. The final bank contains eight anchors.

This construction is a one-time screening heuristic rather than an objective optimized during RL training. Its purpose is to exclude anchors whose perturbations either have little effect on generation or substantially disrupt coherent responses, rather than to identify a unique optimal bank.

\section{Additional Experimental Results}
\label{sec:appendix_results}

\paragraph{Pre-RL Rollout Diversity Visualization.}
\label{appendix:trajectory_diversity}
We analyze rollouts generated from the Qwen2.5-Math-1.5B checkpoint before RL fine-tuning to isolate the direct effect of anchored noise on rollout generation. We represent each correct trajectory by its set of token-level 4-grams and compute pairwise distances as one minus Jaccard similarity. For each of the 28 prompts with at least two correct standard rollouts, we obtain a prompt-specific two-dimensional MDS embedding over the standard and anchor-perturbed trajectories, center and align it using the standard-rollout centroid and principal axis, and normalize it by the standard-rollout RMS radius before pooling the coordinates across prompts. The plotted boundaries are smoothed convex hulls of the resulting points.

\paragraph{Post-RL AIME Rollout Diversity.}
For Fig.~\ref{fig:aime_grop_vs_noir}, we sample 16 responses per problem from the final Qwen2.5-Math-1.5B GRPO and EDEN checkpoints on 60 AIME 2024 and AIME 2025 problems and retain the correct responses. Each response is assigned a deterministic approach signature defined by the set of predefined reasoning markers detected in its complete trace, including summation, multiplication, variable substitution, system solving, exponential reformulation, polynomial division, remainder-based reasoning, case analysis, and modular arithmetic. Correct rollouts are grouped according to the number of distinct signatures observed for their source problem.

\paragraph{Training-Seed Robustness on Qwen2.5-0.5B.}
We repeat GRPO and EDEN with three independent training seeds under the same configuration. Table~\ref{tab:seed_robustness_05b} reports the mean and sample standard deviation. EDEN improves the macro average for all three paired seeds under both greedy and Pass@8 evaluation.

\begin{table}[htp]
\centering
\caption{Training-seed robustness on Qwen2.5-0.5B. Results are reported as mean $\pm$ sample standard deviation over three independent training runs. Pass@1 uses greedy decoding with $T=0$ and top-$p=1.0$; Pass@8 uses $T=0.6$ and top-$p=0.95$.}
\label{tab:seed_robustness_05b}
\small
\setlength{\tabcolsep}{6pt}
\begin{tabular}{llccc}
\toprule
Metric & Method & MATH500 & GSM8K & Avg. \\
\midrule
\multirow{2}{*}{Pass@1}
& GRPO & $31.40 \pm 1.11$ & $59.59 \pm 0.40$ & $45.50 \pm 0.76$ \\
& \modelname{} & $\mathbf{34.80 \pm 1.00}$ & $\mathbf{61.13 \pm 0.42}$ & $\mathbf{47.96 \pm 0.71}$ \\
\midrule
\multirow{2}{*}{Pass@8}
& GRPO & $57.80 \pm 1.25$ & $74.32 \pm 0.56$ & $66.06 \pm 0.90$ \\
& \modelname{} & $\mathbf{59.40 \pm 0.80}$ & $\mathbf{78.60 \pm 0.27}$ & $\mathbf{69.00 \pm 0.53}$ \\
\bottomrule
\end{tabular}
\end{table}

\paragraph{Average Performance Across Rollouts.}
In addition to Pass@$k$, which measures whether at least one of $k$ sampled responses is correct, we report Avg@4 as the mean correctness over four independently sampled responses per prompt. Pass@1 uses greedy decoding ($T=0$), and all trained models are evaluated using their final training checkpoints.

\begin{figure}[htbp]
\centering

\begin{minipage}{0.48\linewidth}
\centering
\small
\setlength{\tabcolsep}{5pt}
\begin{tabular}{lccc}
\toprule
\textbf{Method} & \textbf{MATH} & \textbf{GSM8K} & \textbf{Avg} \\
\midrule
GRPO (rollout 8) & 32.35 & 58.24 & 45.30 \\
GRPO (rollout 16) & 33.10 & 58.70 & 45.90 \\
EDEN (rollout 16) & \textbf{34.10} & \textbf{59.95} & \textbf{47.03} \\
\bottomrule
\end{tabular}
\end{minipage}
\hfill
\begin{minipage}{0.48\linewidth}
\centering
\includegraphics[width=\linewidth]{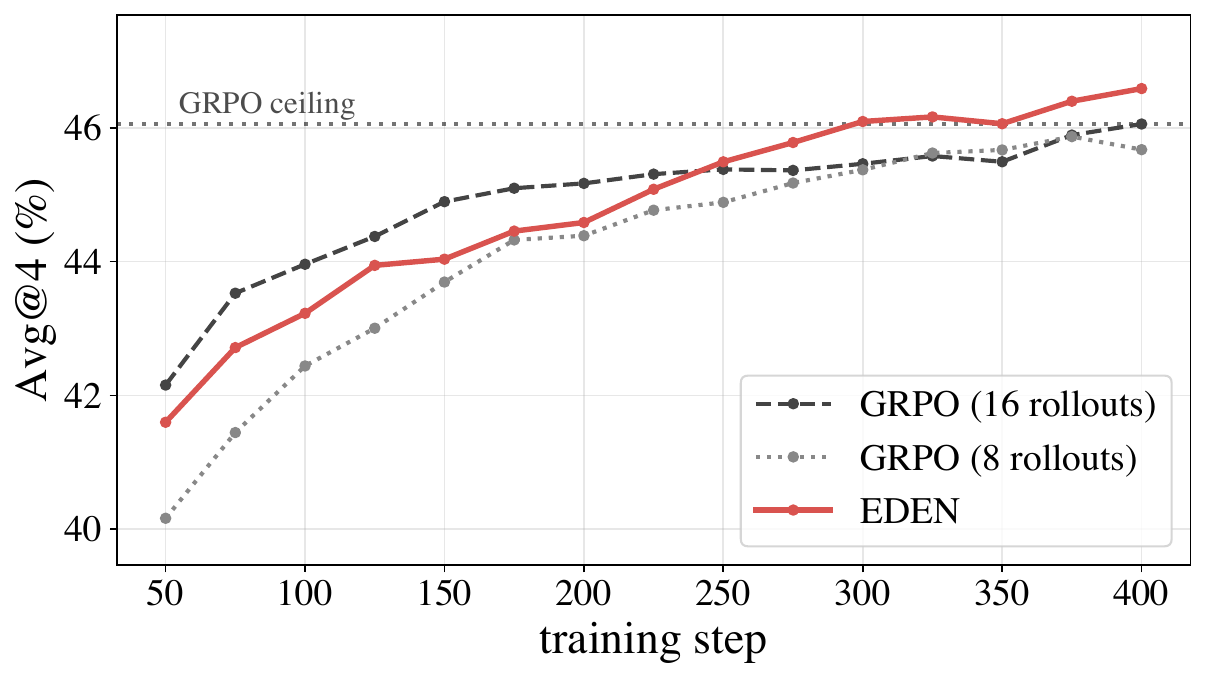}
\end{minipage}

\caption{Average performance (Avg@4) comparison on Qwen2.5-0.5B under $T{=}1$, top-$p{=}0.7$. Left: numerical results. Right: visualization of Avg@4 performance.}
\label{fig:avg4_combined}
\end{figure}

As shown in Fig.~\ref{fig:avg4_combined}, Table ~\ref{tab:avg4_1_5b_math} and Table~\ref{tab:avg4_7b}, EDEN consistently improves Avg@4 across both small-scale and math-specialized models. Notably, these gains are observed under a relatively high sampling temperature ($T=1$), where standard rollouts typically exhibit higher variance.
This suggests that the improvements from EDEN are not limited to occasional high-quality samples, but extend to average correctness across sampled responses.

\begin{table}[htbp]
\centering
\caption{Average performance (Avg@4) comparison on Qwen2.5-Math-1.5B models under $T{=}1$, top-$p{=}0.7$.}
\label{tab:avg4_1_5b_math}
\small
\setlength{\tabcolsep}{4pt}
\begin{tabular}{lccccccccc}
\toprule
\textbf{Method} 
& \textbf{MATH} 
& \textbf{GSM8K} 
& \textbf{AMC} 
& \textbf{AIME24} 
& \textbf{AIME25} 
& \textbf{Olympiad} 
& \textbf{Gaokao} 
& \textbf{Minerva} 
& \textbf{Avg} \\
\midrule
GRPO 
& \textbf{78.00} 
& \textbf{85.58} 
& 56.33 
& \textbf{16.67}
& 6.67 
& 40.47 
& 64.03 
& 31.53 
& 47.41 \\

EDEN
& 77.30 
& 84.95 
& \textbf{57.83} 
& \textbf{16.67} 
& \textbf{11.67} 
& \textbf{41.88} 
& \textbf{66.75} 
& \textbf{32.81} 
& \textbf{48.73} \\
\bottomrule
\end{tabular}
\end{table}

\begin{table}[htbp]
\centering
\caption{Average performance (Avg@4) comparison on Qwen2.5-Math-7B under $T{=}1$, top-$p{=}0.7$.}
\label{tab:avg4_7b}
\small
\setlength{\tabcolsep}{4pt}
\begin{tabular}{lccccccccc}
\toprule
\textbf{Method} 
& \textbf{MATH} 
& \textbf{GSM8K} 
& \textbf{AMC} 
& \textbf{AIME24} 
& \textbf{AIME25} 
& \textbf{Olympiad} 
& \textbf{Gaokao} 
& \textbf{Minerva} 
& \textbf{Avg} \\
\midrule
GRPO 
& 81.65 
& 88.70 
& \textbf{70.18} 
& 29.17 
& \textbf{15.83} 
& \textbf{46.66} 
& 67.99 
& 36.12 
& 54.54 \\

EDEN 
& \textbf{83.25} 
& \textbf{90.09} 
& 65.06 
& \textbf{37.50} 
& 12.50 
& 45.14 
& \textbf{68.77} 
& \textbf{36.67} 
& \textbf{54.87} \\
\bottomrule
\end{tabular}
\end{table}

\section{Extended Ablation Studies}
\label{appendix:ablation_studeis}

\paragraph{Effect of Temperature Schedule.}
We study the role of the cosine temperature schedule in EDEN by analyzing both its necessity and the effect of its duration.

\begin{figure}[!htb]
\centering

\begin{subfigure}{0.48\linewidth}
    \centering
    \includegraphics[width=\linewidth]{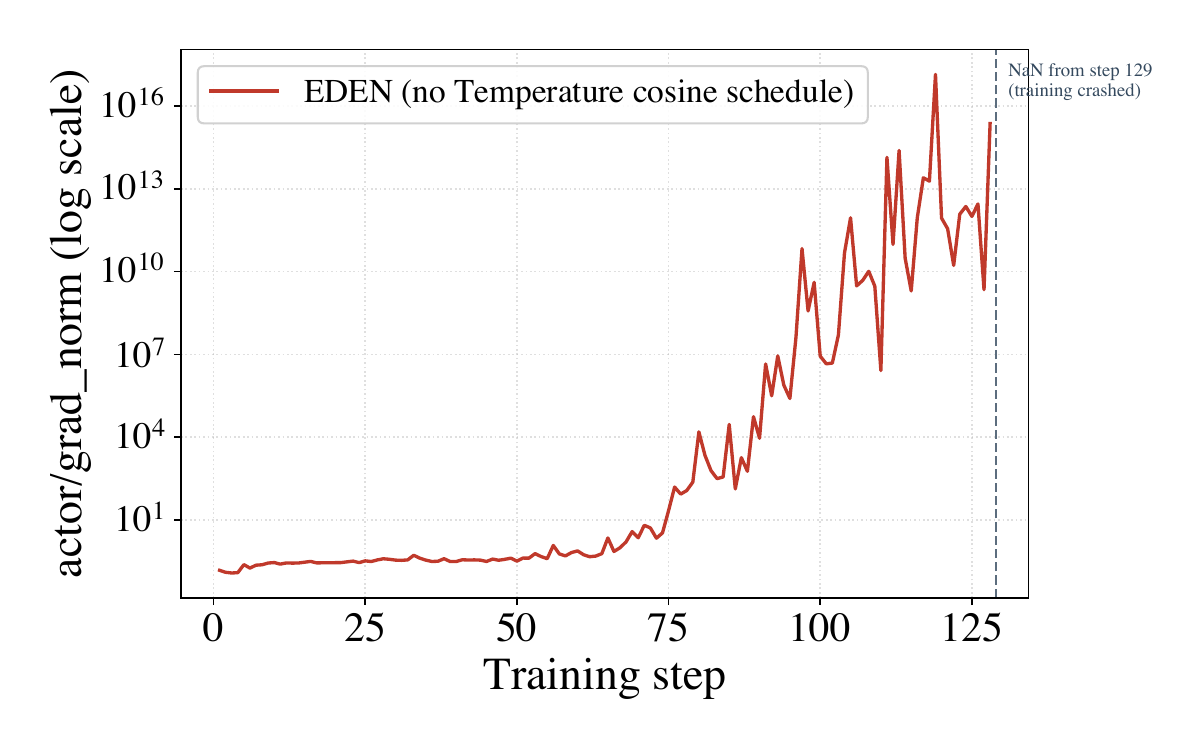}
    \caption{Gradient norm (log scale)}
    \label{fig:grad_norm_no_cos}
\end{subfigure}
\hfill
\begin{subfigure}{0.48\linewidth}
    \centering
    \includegraphics[width=\linewidth]{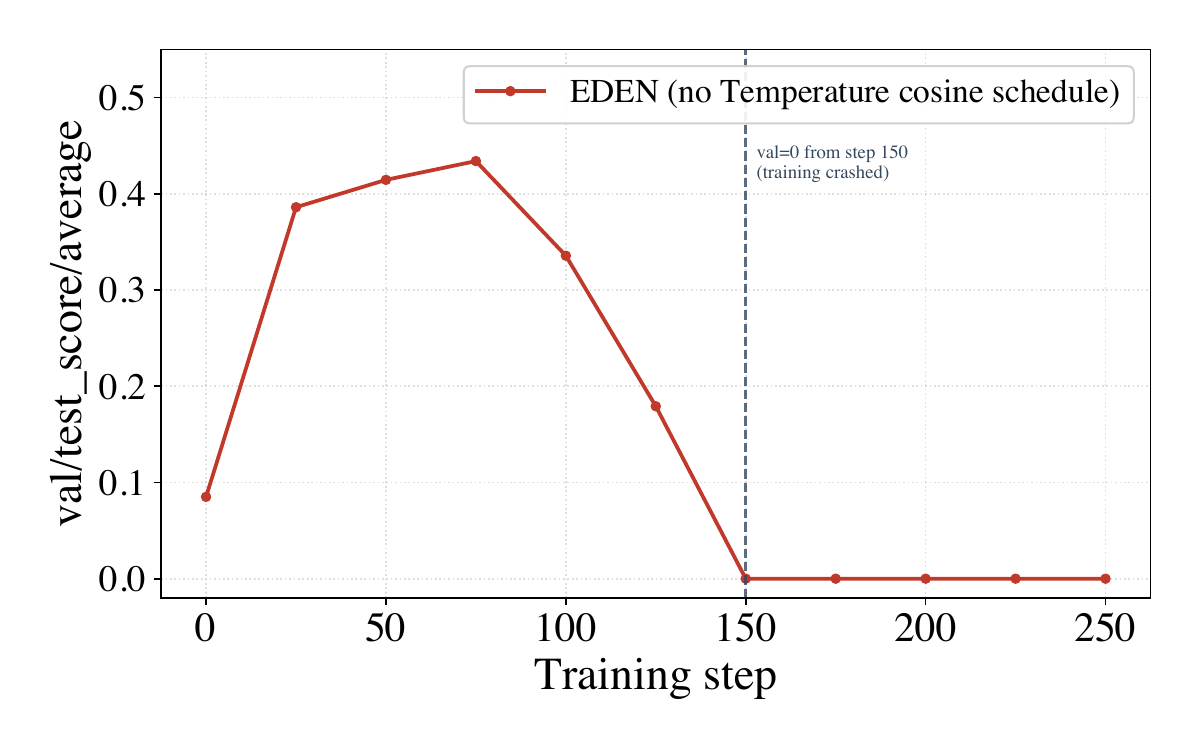}
    \caption{Validation accuracy (Avg@4)}
    \label{fig:validate_no_cos}
\end{subfigure}

\caption{
Training instability without temperature scheduling on Qwen2.5-0.5B. Left: gradient norm (log scale) rapidly increases and eventually diverges, leading to numerical instability (NaNs). Right: validation accuracy initially improves but subsequently collapses to zero as training becomes unstable. These results demonstrate that temperature scheduling is essential for stabilizing noise-perturbed rollouts in EDEN.
}
\label{fig:no_cosine}
\end{figure}

As shown in Fig.~\ref{fig:no_cosine}, removing the temperature schedule leads to unstable training. The gradient norm grows rapidly and eventually diverges, resulting in numerical instability (NaNs), while validation accuracy initially improves but subsequently collapses to zero. This demonstrates that gradually increasing the sampling temperature is essential for stabilizing noise-perturbed rollouts during early training.

\begin{figure}[!htb]
\centering

\begin{subfigure}{0.48\linewidth}
    \centering
    \includegraphics[width=\linewidth]{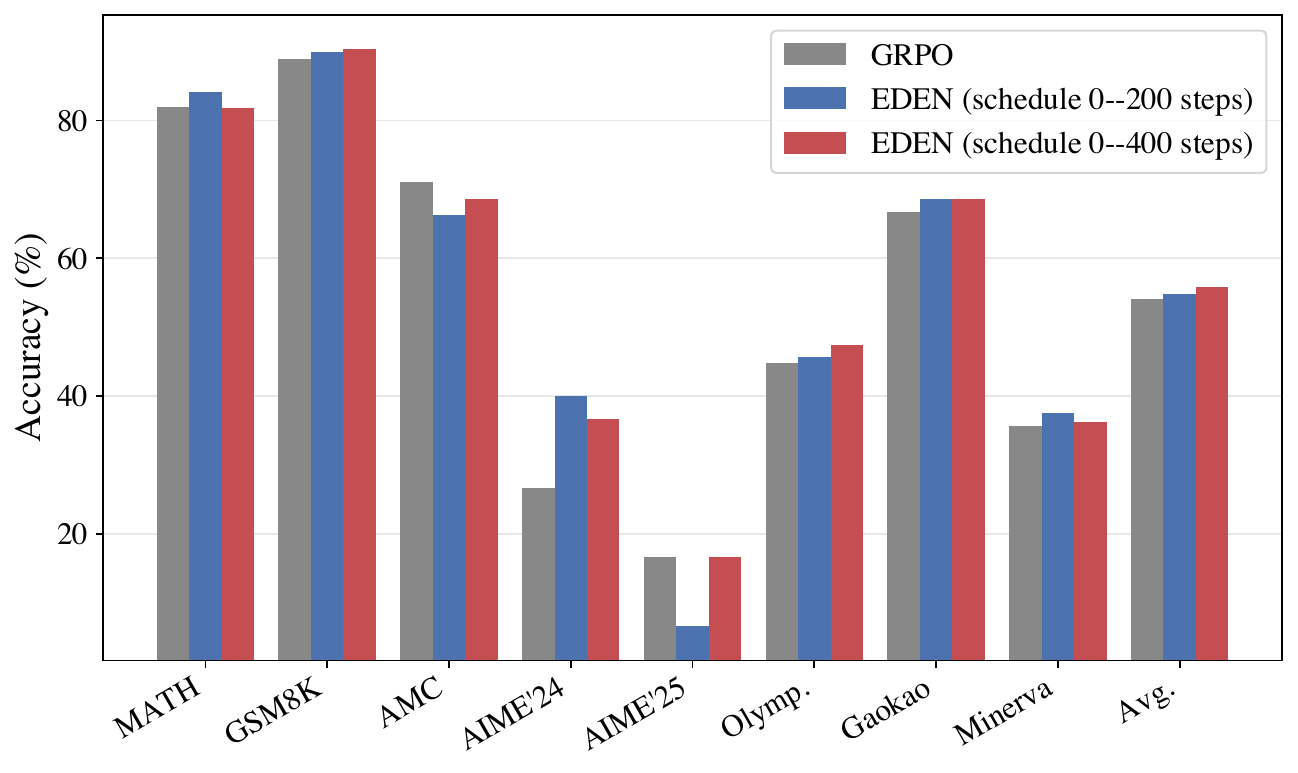}
    \caption{Pass@1 (greedy)}
    \label{fig:schedule_cosine_pass1}
\end{subfigure}
\hfill
\begin{subfigure}{0.48\linewidth}
    \centering
    \includegraphics[width=\linewidth]{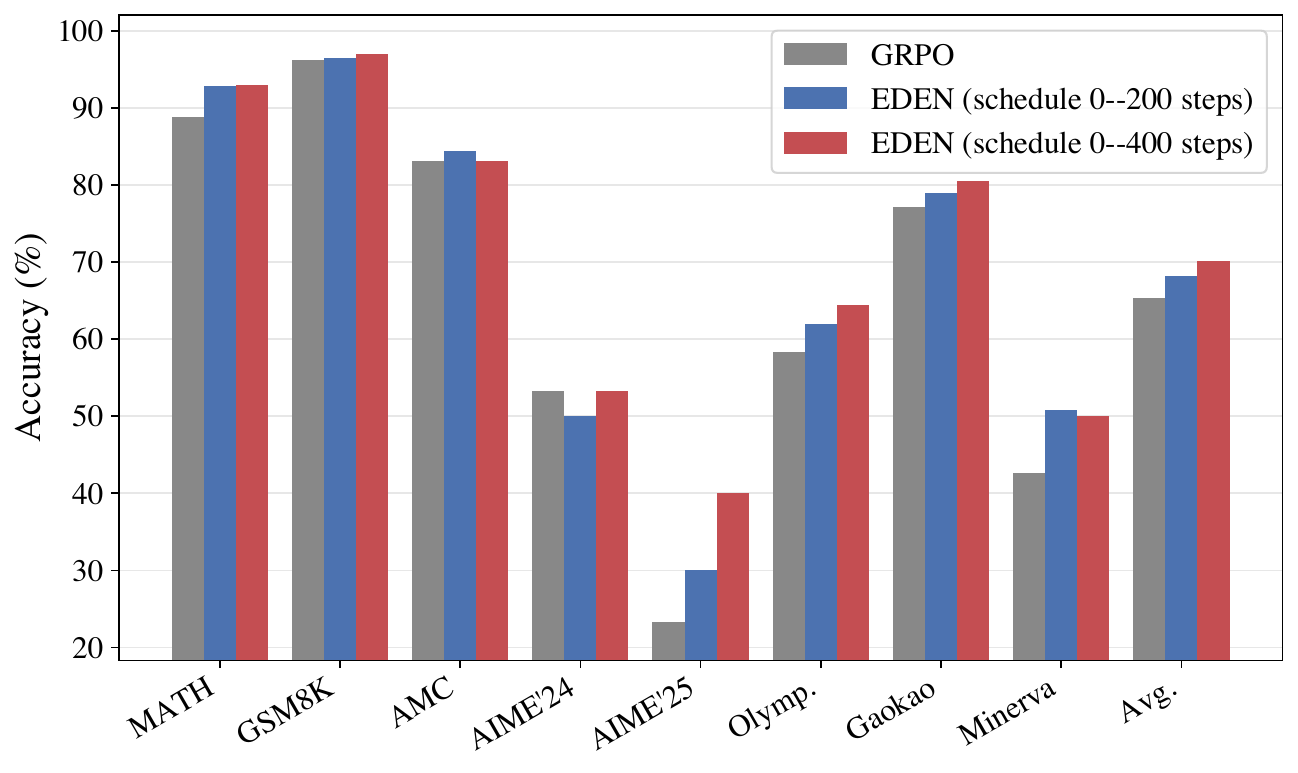}
    \caption{Pass@8 ($T{=}0.6$, top-$p{=}0.95$)}
    \label{fig:schedule_cosine_pass8}
\end{subfigure}

\caption{
Effect of temperature schedule duration in EDEN on Qwen2.5-Math-7B. We compare cosine schedules with durations of 0--200 and 0--400 steps. Both configurations yield comparable performance and consistently outperform the GRPO baseline across most benchmarks. Left: Pass@1 with greedy decoding. Right: Pass@8 with $T{=}0.6$ and top-$p{=}0.95$.
}
\label{fig:schedule_ablation}
\end{figure}

We further investigate the impact of the schedule duration by comparing cosine schedules of 0--200 and 0--400 steps. As shown in Fig.~\ref{fig:schedule_ablation}, both configurations achieve comparable performance across most benchmarks and consistently outperform the GRPO baseline under both Pass@1 (greedy) and Pass@8 (sampling) settings. While extending the schedule to 0--400 steps yields slight improvements on more challenging tasks, the overall differences remain modest.

These results indicate that temperature scheduling is necessary for stable optimization, while performance is relatively insensitive to the schedule duration. This suggests that the primary performance gains of EDEN stem from the proposed noise-based exploration mechanism rather than fine-tuned temperature scheduling.

\paragraph{Effect of Extreme Temperature Sampling.}
We observe that increasing temperature alone does not lead to consistent improvements across benchmarks. While moderate temperature (e.g., $T=1.2$) can improve performance on certain tasks, the gains are unstable and vary significantly across benchmarks.

At extreme temperature ($T=1.4$), training becomes highly unstable under the default setting due to excessively large entropy and stochasticity. To enable a meaningful comparison, we set the entropy coefficient to zero in both GRPO and EDEN, as the exploration induced by high temperature alone is already sufficient. Without this adjustment, training frequently collapses.

\begin{table}[!htbp]
\centering
\caption{Effect of temperature on mathematical reasoning performance (Qwen2.5-Math-1.5B). We compare GRPO and EDEN under different temperature settings. We report Pass@1 (greedy) and Pass@8 (sampling). \textbf{Bold} indicates the best result within each block.}
\label{tab:temperature_effects}
\setlength{\tabcolsep}{4pt}
\renewcommand{\arraystretch}{1.1}
\resizebox{\textwidth}{!}{%
\begin{tabular}{l l cccccccc c}
\toprule
Metric & Method & MATH & GSM8K & AMC & AIME'24 & AIME'25 & Olympiad & Gaokao & Minerva & Avg. \\
\midrule

\multirow{5}{*}{Pass@1} 
 & GRPO ($T{=}1$)       & 76.60 & \textbf{86.35} & 53.01 & 13.33 & \textbf{20.00} & 40.50 & 64.94 & 31.99 & 48.34 \\
 & GRPO ($T{=}1.2$)     & 77.00 & 85.29 & 56.63 & 13.33 & 10.00 & 39.02 & 65.71 & 30.15 & 47.14 \\
 & GRPO ($T{=}1.4$)     & 76.60 & 84.91 & 59.04 & 13.33 & 16.67 & 41.10 & 66.49 & 31.99 & 48.77 \\
 & EDEN ($T{=}1$)       & 77.40 & 86.28 & \textbf{63.86} & 13.33 & 13.33 & \textbf{41.84} & \textbf{67.01} & 31.62 & \textbf{49.33} \\
 & EDEN ($T{=}1.4$)     & \textbf{78.60} & 85.75 & 57.83 & \textbf{20.00} & 13.33 & 39.17 & 64.94 & \textbf{33.09} & 49.09 \\

\midrule

\multirow{5}{*}{Pass@8} 
 & GRPO ($T{=}1$)       & 86.60 & 94.92 & 69.88 & \textbf{33.33} & 16.67 & 55.34 & 77.14 & 41.54 & 59.43 \\
 & GRPO ($T{=}1.2$)     & 88.20 & 91.89 & 71.08 & 26.67 & \textbf{33.33} & 53.26 & 77.40 & 42.28 & 60.51 \\
 & GRPO ($T{=}1.4$)     & 87.80 & 91.74 & 75.90 & \textbf{33.33} & 26.67 & 54.75 & 75.58 & 41.18 & 60.87 \\
 & EDEN ($T{=}1$)       & \textbf{89.60} & \textbf{95.60} & \textbf{79.52} & \textbf{33.33} & 20.00 & \textbf{55.93} & \textbf{78.96} & \textbf{44.12} & \textbf{62.13} \\
 & EDEN ($T{=}1.4$)     & 87.80 & 94.77 & 71.08 & 30.00 & 20.00 & 55.79 & 75.58 & 42.28 & 59.66 \\

\bottomrule
\end{tabular}%
}
\end{table}

Under this controlled setting, we find that extreme temperature does not consistently improve performance and often leads to degradation, particularly for Pass@8. In contrast, EDEN achieves the best overall performance at $T=1$, suggesting that structured noise-based exploration provides more effective and stable exploration than simply increasing sampling temperature.

\section{Case Studies.}
\label{appendix:case_studies}
We present a case study on AIME to examine how post-training rollout diversity appears in complete reasoning traces. For a selected problem, we compare correct rollouts generated by GRPO and EDEN from 16 samples per prompt using $T{=}0.6$ and top-$p{=}0.95$. These examples provide qualitative context for the aggregate AIME rollout diversity in Fig.~\ref{fig:aime_grop_vs_noir}. Here, ``correct'' refers to final-answer correctness, not the validity of every intermediate reasoning step.

\paragraph{Problem.}
Find the sum of all positive integers $ n $ such that $ n + 2 $ divides the product $ 3(n + 3)(n^2 + 9) $
\paragraph{Correct answer.} $49$
\paragraph{Analysis of Reasoning Traces.}
We compare how the correct rollouts formulate the underlying divisibility argument.

For GRPO, the two correct solutions follow essentially the same derivation. Both introduce the substitution $k=n+2$, rewrite the expression in terms of $k$, and reduce the divisibility condition to $k\mid39$ through algebraic simplification and coprimality arguments. Their differences are largely limited to the order of expansion and intermediate steps.

EDEN, by contrast, produces a wider range of derivations: (i) polynomial division by $(n+2)$ to obtain the constant remainder $39$; (ii) parameterization via $n=kx-2$ followed by modular simplification; (iii) direct use of the remainder theorem through $n\equiv-2\pmod{n+2}$; and (iv) substitution $k=n+2$ followed by algebraic expansion and modular reduction. Although these derivations share the same core reduction $n+2\mid39$, they reach it through different algebraic formulations and sequences of intermediate steps.

This example illustrates that EDEN produces greater variation in complete reasoning traces than GRPO, spanning several derivations of the same core divisibility argument. This qualitative pattern complements the aggregate post-training rollout diversity shown in Fig.~\ref{fig:aime_grop_vs_noir}.

\paragraph{EDEN --- Rollout 1 (correct, $\checkmark$).}
\begin{quote}\small\itshape
We need to find all positive integers \( n \) such that \( n + 2 \) divides the product \( 3(n + 3)(n^2 + 9) \). This means that \( \frac{3(n + 3)(n^2 + 9)}{n + 2} \) must be an integer.

First, let's perform polynomial long division on \( \frac{3(n + 3)(n^2 + 9)}{n + 2} \).

We start by expanding the numerator:
\[ 3(n + 3)(n^2 + 9) = 3(n^3 + 9n + 3n^2 + 27) = 3(n^3 + 3n^2 + 9n + 27) = 3n^3 + 9n^2 + 27n + 81. \]

Now, we need to divide \( 3n^3 + 9n^2 + 27n + 81 \) by \( n + 2 \). Using polynomial long division:

1. Divide the leading term \( 3n^3 \) by \( n \) to get \( 3n^2 \).
2. Multiply \( 3n^2 \) by \( n + 2 \) to get \( 3n^3 + 6n^2 \).
3. Subtract \( 3n^3 + 6n^2 \) from \( 3n^3 + 9n^2 + 27n + 81 \) to get \( 3n^2 + 27n + 81 \).
4. Divide the leading term \( 3n^2 \) by \( n \) to get \( 3n \).
5. Multiply \( 3n \) by \( n + 2 \) to get \( 3n^2 + 6n \).
6. Subtract \( 3n^2 + 6n \) from \( 3n^2 + 27n + 81 \) to get \( 21n + 81 \).
7. Divide the leading term \( 21n \) by \( n \) to get \( 21 \).
8. Multiply \( 21 \) by \( n + 2 \) to get \( 21n + 42 \).
9. Subtract \( 21n + 42 \) from \( 21n + 81 \) to get \( 39 \).

So, we have:
\[ \frac{3n^3 + 9n^2 + 27n + 81}{n + 2} = 3n^2 + 3n + 21 + \frac{39}{n + 2}. \]

For \( \frac{3(n + 3)(n^2 + 9)}{n + 2} \) to be an integer, \( \frac{39}{n + 2} \) must be an integer. This means \( n + 2 \) must be a divisor of 39. The divisors of 39 are \( \pm 1, \pm 3, \pm 13, \pm 39 \). Since \( n \) is a positive integer, \( n + 2 \) must be one of the positive divisors of 39, i.e., \( 1, 3, 13, 39 \).

So, we solve for \( n \):
1. \( n + 2 = 1 \) gives \( n = -1 \) (not a positive integer).
2. \( n + 2 = 3 \) gives \( n = 1 \).
3. \( n + 2 = 13 \) gives \( n = 11 \).
4. \( n + 2 = 39 \) gives \( n = 37 \).

Thus, the possible values for \( n \) are \( 1, 11, \) and \( 37 \). The sum of these values is:
\[ 1 + 11 + 37 = 49. \]

So, the sum of all positive integers \( n \) such that \( n + 2 \) divides \( 3(n + 3)(n^2 + 9) \) is \(\boxed{49}\).
\end{quote}

\paragraph{EDEN --- Rollout 2 (correct, $\checkmark$).}
\begin{quote}\small\itshape
We need to find all positive integers \( n \) such that \( n + 2 \) divides the product \( 3(n + 3)(n^2 + 9) \). This means that:
\[ 3(n + 3)(n^2 + 9) \equiv 0 \pmod{n + 2}. \]

To simplify this, we will use the fact that if \( n + 2 \equiv 0 \pmod{k} \), then \( n \equiv -2 \pmod{k} \), or \( n = kx - 2 \) for some integer \( x \). Substituting \( n = kx - 2 \) into the expression \( 3(n + 3)(n^2 + 9) \), we get:
\[ n + 3 = (kx - 2) + 3 = kx + 1, \]
and
\[ n^2 + 9 = (kx - 2)^2 + 9 = k^2x^2 - 4kx + 4 + 9 = k^2x^2 - 4kx + 13. \]
So the product becomes:
\[ 3(n + 3)(n^2 + 9) = 3(kx + 1)(k^2x^2 - 4kx + 13). \]
We need this to be divisible by \( k = n + 2 \). Therefore:
\[ 3(kx + 1)(k^2x^2 - 4kx + 13) \equiv 0 \pmod{k}. \]
Since \( k \) divides \( k^2x^2 \) and \( 4kx \), it must also divide \( kx + 1 \) and \( k^2x^2 - 4kx + 13 \). So:
\[ kx + 1 \equiv 1 \pmod{k}, \]
and
\[ k^2x^2 - 4kx + 13 \equiv 13 \pmod{k}. \]
Thus, we need:
\[ 3 \cdot 1 \cdot 13 \equiv 0 \pmod{k}, \]
which simplifies to:
\[ 39 \equiv 0 \pmod{k}. \]
This means \( k \) must be a divisor of 39. The divisors of 39 are \( 1, 3, 13, 39 \). We will check each case:

1. \( k = 1 \):
\[ n + 2 = 1 \Rightarrow n = -1, \]
which is not a positive integer.

2. \( k = 3 \):
\[ n + 2 = 3 \Rightarrow n = 1. \]
We substitute \( n = 1 \) into the original expression:
\[ 3(n + 3)(n^2 + 9) = 3(1 + 3)(1^2 + 9) = 3 \cdot 4 \cdot 10 = 120. \]
And \( 120 \div 3 = 40 \), which is an integer. So \( n = 1 \) works.

3. \( k = 13 \):
\[ n + 2 = 13 \Rightarrow n = 11. \]
We substitute \( n = 11 \) into the original expression:
\[ 3(n + 3)(n^2 + 9) = 3(11 + 3)(11^2 + 9) = 3 \cdot 14 \cdot (121 + 9) = 3 \cdot 14 \cdot 130 = 5460. \]
And \( 5460 \div 13 = 420 \), which is an integer. So \( n = 11 \) works.

4. \( k = 39 \):
\[ n + 2 = 39 \Rightarrow n = 37. \]
We substitute \( n = 37 \) into the original expression:
\[ 3(n + 3)(n^2 + 9) = 3(37 + 3)(37^2 + 9) = 3 \cdot 40 \cdot (1369 + 9) = 3 \cdot 40 \cdot 1378 = 165360. \]
And \( 165360 \div 39 = 4240 \), which is an integer. So \( n = 37 \) works.

So the positive integers \( n \) that satisfy the condition are \( 1, 11, \) and \( 37 \). The sum of these integers is:
\[ 1 + 11 + 37 = 49. \]

Thus, the final answer is:
\[
\boxed{49}
\]
\end{quote}

\paragraph{EDEN --- Rollout 3 (correct, $\checkmark$).}
\begin{quote}\small\itshape
We need to find all positive integers \( n \) such that \( n + 2 \) divides the product \( 3(n + 3)(n^2 + 9) \). This means that:
\[ 3(n + 3)(n^2 + 9) \equiv 0 \pmod{n + 2} \]

To simplify this, we will use the fact that if \( n + 2 \mid k \), then \( n \equiv -2 \pmod{n + 2} \). So, we substitute \( n = -2 \) into the expression \( 3(n + 3)(n^2 + 9) \):
\[ 3((-2) + 3)((-2)^2 + 9) = 3(1)(4 + 9) = 3 \cdot 1 \cdot 13 = 39. \]
This means that \( n + 2 \) must divide 39. The divisors of 39 are \( \pm 1, \pm 3, \pm 13, \pm 39 \). Since \( n \) is a positive integer, \( n + 2 \) must be one of the positive divisors of 39, i.e., \( 1, 3, 13, 39 \). Therefore, we have:
\[ n + 2 = 1 \implies n = -1 \quad \text{(not a positive integer)}, \]
\[ n + 2 = 3 \implies n = 1, \]
\[ n + 2 = 13 \implies n = 11, \]
\[ n + 2 = 39 \implies n = 37. \]

So, the possible values for \( n \) are \( 1, 11, \) and \( 37 \). Now, we need to check if these values satisfy the original condition:
1. For \( n = 1 \):
\[ n + 2 = 3, \]
\[ 3(n + 3)(n^2 + 9) = 3(1 + 3)(1^2 + 9) = 3 \cdot 4 \cdot 10 = 120. \]
Since \( 3 \mid 120 \), \( n = 1 \) is a valid solution.
2. For \( n = 11 \):
\[ n + 2 = 13, \]
\[ 3(n + 3)(n^2 + 9) = 3(11 + 3)(11^2 + 9) = 3 \cdot 14 \cdot (121 + 9) = 3 \cdot 14 \cdot 130 = 5460. \]
Since \( 13 \mid 5460 \), \( n = 11 \) is a valid solution.
3. For \( n = 37 \):
\[ n + 2 = 39, \]
\[ 3(n + 3)(n^2 + 9) = 3(37 + 3)(37^2 + 9) = 3 \cdot 40 \cdot (1369 + 9) = 3 \cdot 40 \cdot 1378 = 165360. \]
Since \( 39 \mid 165360 \), \( n = 37 \) is a valid solution.

So, the valid values for \( n \) are \( 1, 11, \) and \( 37 \). The sum of these values is:
\[ 1 + 11 + 37 = 49. \]

Thus, the sum of all positive integers \( n \) such that \( n + 2 \) divides the product \( 3(n + 3)(n^2 + 9) \) is \(\boxed{49}\).
\end{quote}

\paragraph{EDEN --- Rollout 4 (correct, $\checkmark$).}
\begin{quote}\small\itshape
We need to find all positive integers \( n \) such that \( n + 2 \) divides the product \( 3(n + 3)(n^2 + 9) \). This means that \( 3(n + 3)(n^2 + 9) \equiv 0 \pmod{n + 2} \).

First, let's substitute \( n = k - 2 \) to simplify the expression. We need to find \( k \) such that \( k \) divides \( 3(k + 1)((k-2)^2 + 9) \).

So, we substitute \( n = k - 2 \):
\[ 3(n + 3)(n^2 + 9) = 3((k-2) + 3)((k-2)^2 + 9) = 3(k + 1)((k-2)^2 + 9). \]

Now, we simplify \((k-2)^2 + 9\):
\[ (k-2)^2 + 9 = k^2 - 4k + 4 + 9 = k^2 - 4k + 13. \]

So, we need \( k \) to divide:
\[ 3(k + 1)(k^2 - 4k + 13). \]

We need to find the values of \( k \) such that \( k \) divides \( 3(k + 1)(k^2 - 4k + 13) \). This means:
\[ 3(k + 1)(k^2 - 4k + 13) \equiv 0 \pmod{k}. \]

Since \( k \) divides \( k^2 \) and \( 4k \), we have:
\[ 3(k + 1)(k^2 - 4k + 13) \equiv 3(k + 1)(13) \equiv 39(k + 1) \pmod{k}. \]

This simplifies to:
\[ 39(k + 1) \equiv 0 \pmod{k}. \]

This means \( k \) must divide \( 39(k + 1) \). Since \( k \) divides \( 39(k + 1) \), it must divide \( 39(k + 1) - 39k = 39 \). So, \( k \) must be a divisor of 39. The divisors of 39 are \( 1, 3, 13, 39 \).

We now need to check each of these values for \( k \):

1. If \( k = 1 \), then \( n = k - 2 = 1 - 2 = -1 \), which is not a positive integer.
2. If \( k = 3 \), then \( n = k - 2 = 3 - 2 = 1 \). We check:
   \[ 3(n + 3)(n^2 + 9) = 3(1 + 3)(1^2 + 9) = 3 \cdot 4 \cdot 10 = 120. \]
   And \( n + 2 = 1 + 2 = 3 \), which divides 120. So \( n = 1 \) works.
3. If \( k = 13 \), then \( n = k - 2 = 13 - 2 = 11 \). We check:
   \[ 3(n + 3)(n^2 + 9) = 3(11 + 3)(11^2 + 9) = 3 \cdot 14 \cdot (121 + 9) = 3 \cdot 14 \cdot 130 = 5460. \]
   And \( n + 2 = 11 + 2 = 13 \), which divides 5460. So \( n = 11 \) works.
4. If \( k = 39 \), then \( n = k - 2 = 39 - 2 = 37 \). We check:
   \[ 3(n + 3)(n^2 + 9) = 3(37 + 3)(37^2 + 9) = 3 \cdot 40 \cdot (1369 + 9) = 3 \cdot 40 \cdot 1378 = 165360. \]
   And \( n + 2 = 37 + 2 = 39 \), which divides 165360. So \( n = 37 \) works.

So, the positive integers \( n \) that satisfy the condition are \( 1, 11, \) and \( 37 \). The sum of these integers is:
\[ 1 + 11 + 37 = 49. \]

Therefore, the final answer is:
\[
\boxed{49}
\]
\end{quote}

\paragraph{GRPO Baseline --- Rollout 1 (correct, $\checkmark$).}
\begin{quote}\small\itshape
To solve the problem, we need to find all positive integers \( n \) such that \( n + 2 \) divides the product \( 3(n + 3)(n^2 + 9) \).

Let's denote \( k = n + 2 \). Then \( n = k - 2 \). We need to find \( n \) such that \( k \) divides \( 3(k - 2 + 3)((k - 2)^2 + 9) \). Simplifying the expression inside the divisors, we get:
\[ k \div 3(k - 2 + 3)((k - 2)^2 + 9) = k \div 3(k + 1)((k - 2)^2 + 9). \]
Now, let's simplify \((k - 2)^2 + 9\):
\[ (k - 2)^2 + 9 = k^2 - 4k + 4 + 9 = k^2 - 4k + 13. \]
So, the condition becomes:
\[ k \div 3(k + 1)(k^2 - 4k + 13). \]
This means that \( 3(k + 1)(k^2 - 4k + 13) \) must be divisible by \( k \). Therefore, we need:
\[ 3(k + 1)(k^2 - 4k + 13) \equiv 0 \pmod{k}. \]
Since \( k \) divides \( k(k^2 - 4k + 13) \), we only need to check the divisibility by 3:
\[ 3(k + 1)(k^2 - 4k + 13) \equiv 0 \pmod{k} \implies 3(k + 1) \left( k^2 - 4k + 13 \right) \equiv 0 \pmod{k}. \]
This simplifies to checking the divisibility of \( 3(k + 1) \left( k^2 - 4k + 13 \right) \) by \( k \). Since \( k \) and \( k + 1 \) are consecutive integers, they are coprime, so \( k \) must divide \( 3(k^2 - 4k + 13) \). Therefore, we need:
\[ 3(k^2 - 4k + 13) \equiv 0 \pmod{k}. \]
This further simplifies to:
\[ 3(k^2 - 4k + 13) = 3k^2 - 12k + 39. \]
We need:
\[ 3k^2 - 12k + 39 \equiv 0 \pmod{k}. \]
Since \( 3k^2 \) and \( -12k \) are both divisible by \( k \), we are left with:
\[ 39 \equiv 0 \pmod{k}. \]
This means that \( k \) must be a divisor of 39. The divisors of 39 are \( 1, 3, 13, 39 \). However, since \( k = n + 2 \), \( k \) must be greater than 2. Therefore, the possible values for \( k \) are \( 3, 13, 39 \).

Now, we can find the corresponding values of \( n \):
1. If \( k = 3 \), then \( n = k - 2 = 3 - 2 = 1 \).
2. If \( k = 13 \), then \( n = k - 2 = 13 - 2 = 11 \).
3. If \( k = 39 \), then \( n = k - 2 = 39 - 2 = 37 \).

The positive integers \( n \) that satisfy the condition are \( 1, 11, \) and \( 37 \). The sum of these integers is:
\[ 1 + 11 + 37 = 49. \]

Therefore, the sum of all positive integers \( n \) such that \( n + 2 \) divides the product \( 3(n + 3)(n^2 + 9) \) is \(\boxed{49}\).
\end{quote}

\paragraph{GRPO Baseline --- Rollout 2 (correct, $\checkmark$).}
\begin{quote}\small\itshape
To solve the problem, we need to find all positive integers \( n \) such that \( n + 2 \) divides the product \( 3(n + 3)(n^2 + 9) \).

Let's denote \( k = n + 2 \). Then \( n = k - 2 \). We need to find \( n \) such that \( k \) divides \( 3(k - 2 + 3)((k - 2)^2 + 9) \). Simplifying the expression inside the divisors, we get:
\[ k \div 3(k - 2 + 3)((k - 2)^2 + 9) = k \div 3(k + 1)((k - 2)^2 + 9). \]
Expanding \((k - 2)^2 + 9\), we get:
\[ (k - 2)^2 + 9 = k^2 - 4k + 4 + 9 = k^2 - 4k + 13. \]
So, the condition becomes:
\[ k \div 3(k + 1)(k^2 - 4k + 13). \]
This means that \( 3(k + 1)(k^2 - 4k + 13) \) must be divisible by \( k \). Therefore, we need:
\[ 3(k + 1)(k^2 - 4k + 13) \equiv 0 \pmod{k}. \]
Since \( k \) divides \( k(k^2 - 4k + 13) \), we only need to check the divisibility by 3:
\[ 3(k + 1)(k^2 - 4k + 13) \equiv 0 \pmod{k} \implies 3(k + 1) \left( k^2 - 4k + 13 \right) \equiv 0 \pmod{k}. \]
This simplifies to checking the divisibility of \( 3(k + 1) \left( k^2 - 4k + 13 \right) \) by \( k \). Since \( k \) and \( k + 1 \) are coprime, \( k \) must divide \( 3(k^2 - 4k + 13) \). So, we need:
\[ 3(k^2 - 4k + 13) \equiv 0 \pmod{k}. \]
This further simplifies to:
\[ 3(k^2 - 4k + 13) = 3k^2 - 12k + 39. \]
We need:
\[ 3k^2 - 12k + 39 \equiv 0 \pmod{k}. \]
Since \( 3k^2 \) and \( -12k \) are both divisible by \( k \), we are left with:
\[ 39 \equiv 0 \pmod{k}. \]
This means that \( k \) must be a divisor of 39. The divisors of 39 are \( 1, 3, 13, 39 \). However, since \( k = n + 2 \), \( k \) must be greater than 2. Therefore, the possible values for \( k \) are \( 3, 13, 39 \).

Now, we can find the corresponding values of \( n \):
1. If \( k = 3 \), then \( n = k - 2 = 3 - 2 = 1 \).
2. If \( k = 13 \), then \( n = k - 2 = 13 - 2 = 11 \).
3. If \( k = 39 \), then \( n = k - 2 = 39 - 2 = 37 \).

So, the positive integers \( n \) that satisfy the condition are \( 1, 11, \) and \( 37 \). The sum of these integers is:
\[ 1 + 11 + 37 = 49. \]

Therefore, the sum of all positive integers \( n \) such that \( n + 2 \) divides the product \( 3(n + 3)(n^2 + 9) \) is \(\boxed{49}\).
\end{quote}

\end{document}

%% file: math_commands.tex
\usepackage{amsmath,amsfonts,bm}

\def\eqref#1{equation~\ref{#1}}
\def\1{\bm{1}}

\DeclareMathAlphabet{\mathsfit}{\encodingdefault}{\sfdefault}{m}{sl}
\SetMathAlphabet{\mathsfit}{bold}{\encodingdefault}{\sfdefault}{bx}{n}